\documentclass{article}

\usepackage[final]{nips_2017}

\usepackage[utf8]{inputenc} % allow utf-8 input
\usepackage[T1]{fontenc}    % use 8-bit T1 fonts
\usepackage{hyperref}       % hyperlinks
\usepackage{url}            % simple URL typesetting
\usepackage{booktabs}       % professional-quality tables
\usepackage{amsfonts}       % blackboard math symbols
\usepackage{nicefrac}       % compact symbols for 1/2, etc.
\usepackage{microtype}      % microtypography
\usepackage{amsmath}
\usepackage{amssymb}
\usepackage{graphicx}
\usepackage{ifthen}
\usepackage[utf8]{inputenc}
\usepackage{color}
\usepackage{amsmath,amsthm}
\usepackage{enumitem}
\usepackage{algorithm,algorithmicx,algcompatible,algpseudocode}
\usepackage[font=small,labelfont=it]{caption}

\definecolor{mydarkblue}{rgb}{0,0.08,0.45}
\hypersetup{
  colorlinks=true,
  linkcolor=mydarkblue,
  citecolor=mydarkblue,
  filecolor=mydarkblue,
  urlcolor=mydarkblue,
}
\setcitestyle{authoryear,round,citesep={;},aysep={,},yysep={;}}

\newcommand\sD{\ensuremath{\mathcal{D}}}

\newcommand\sF{\ensuremath{\mathcal{F}}}

\newcommand\sX{\ensuremath{\mathcal{X}}}
\newcommand\sY{\ensuremath{\mathcal{Y}}}

\newcommand\bE{\ensuremath{\mathbf{E}}}

\newcommand\bL{\ensuremath{\mathbf{L}}}
\newcommand\bM{\ensuremath{\mathbf{M}}}

\newcommand\bR{\ensuremath{\mathbf{R}}}

\newcommand\bZ{\ensuremath{\mathbf{Z}}}

\newcommand\p[1]{\ensuremath{\left( #1 \right)}} % Parenthesis ()
\newcommand\R{\ensuremath{\mathbb{R}}} % Real numbers
\newcommand\eqdef{\ensuremath{\stackrel{\rm def}{=}}} % Equal by definition
\newcommand\refsec[1]{Section~\ref{sec:#1}}

\newcommand\reffig[1]{Figure~\ref{fig:#1}}

\ifthenelse{\isundefined{\definition}}{\newtheorem{definition}{Definition}}{}
\ifthenelse{\isundefined{\assumption}}{\newtheorem{assumption}{Assumption}}{}
\ifthenelse{\isundefined{\hypothesis}}{\newtheorem{hypothesis}{Hypothesis}}{}
\ifthenelse{\isundefined{\proposition}}{\newtheorem{proposition}{Proposition}}{}
\ifthenelse{\isundefined{\theorem}}{\newtheorem{theorem}{Theorem}}{}
\ifthenelse{\isundefined{\lemma}}{\newtheorem{lemma}{Lemma}}{}
\ifthenelse{\isundefined{\corollary}}{\newtheorem{corollary}{Corollary}}{}
\ifthenelse{\isundefined{\alg}}{\newtheorem{alg}{Algorithm}}{}
\ifthenelse{\isundefined{\example}}{\newtheorem{example}{Example}}{}
\newcommand{\E}{\ensuremath{\mathbb{E}}} % Expectation
\newcommand\mnist{MNIST-1-7}

\newcommand\sDc{\sD_\text{c}}
\newcommand\sDp{\sD_\text{p}}
\newcommand\pip{\pi_\text{p}}
\newcommand\sDcp{\sDc \cup \sDp}

\newcommand{\Up}{\tilde{U}}

\newcommand\sFsphere{\sF_\text{sphere}}
\newcommand\sFslab{\sF_\text{slab}}

\DeclareMathOperator*{\argmax}{argmax}
\DeclareMathOperator*{\argmin}{argmin}
\newcommand{\support}{\operatorname{supp}}

\newcommand{\thetaNom}{\hat{\theta}}
\newcommand{\thetaTil}{\tilde{\theta}}
\newcommand{\oo}{\mathcal{O}}

\title{Certified Defenses for Data Poisoning Attacks}

\author{
  Jacob Steinhardt\thanks{Equal contribution.} \\
  Stanford University \\
  {\tt jsteinha@stanford.edu}
\And
  Pang Wei Koh\footnotemark[\value{footnote}] \\
  Stanford University \\
  {\tt pangwei@cs.stanford.edu}
\And
	Percy Liang \\
  Stanford University \\
  {\tt pliang@cs.stanford.edu}
}

\renewcommand{\paragraph}[1]{\textbf{#1}}

\begin{document}

\maketitle

\begin{abstract}
%For machine learning systems trained from user-provided data such as product reviews,
%\pl{don't need to mention 'product reviews' - too specific}
%a non-traditional security concern known as \emph{data poisoning} arises in which malicious
%users can insert entries into the dataset with the aim of corrupting the learned model.
%\pl{can shorten first sentence}
%While recent work has demonstrated a number of data poisoning attacks and defenses,
%little is understood about the fundamental limits of attackability for a given dataset,
%or of the worst-case performance of a defense in the face of a determined attacker.
%In this paper we remedy this by constructing \emph{worst-case upper bounds} on the
%loss of a defender across a wide range of attacks, which come paired with a candidate
%attack that nearly realizes the upper bound. This gives us a powerful tool for quickly
%and automatically assessing defense strategies on a new dataset, which we use to
%empirically investigate which datasets tend to be more or less susceptible to data
%poisoning attacks. Our approach makes use of duality results for no-regret online
%learners, which allow us to quickly compute an approximate Nash equilibrium between
%the attacker and defender.
%\pl{should somehow shorten everything a bit and add a sentence on experimental results:
%we tested on 4 datasets, and found that some datasets are safe}
%
Machine learning systems trained on user-provided data
are susceptible to \emph{data poisoning} attacks, whereby malicious users
inject false training data with the aim of corrupting the learned model.
While recent work has
proposed a number of attacks and defenses, little is understood
about the worst-case loss of a defense in the face of a determined attacker.
We address this by constructing approximate
upper bounds on the loss across a broad family 
of attacks, for defenders that first perform outlier removal followed by empirical risk minimization.
Our approximation relies on two assumptions: (1) that the dataset is large enough for 
statistical concentration between train and test error to hold, and (2) that outliers 
within the clean (non-poisoned) data do not have a strong effect on the model.
Our bound comes paired with a candidate attack
that often nearly matches the upper bound, giving us a powerful tool for quickly
assessing defenses on a given dataset.
Empirically,
we find that even under a simple defense,
the \mnist\ and Dogfish datasets are resilient to attack,
while in contrast the IMDB sentiment dataset can be driven from $12\%$ to
$23\%$ test error by adding only $3\%$ poisoned data.

%\pw{Should we mention specifics/limitations -- classification, convexity, SVMs, ...?}
% PL: probably no room

\end{abstract}

\setcounter{footnote}{0}

\section{Introduction}

Traditionally, computer security seeks to ensure a system's integrity against attackers 
by creating clear boundaries between the system and the outside world \citep{bishop2002art}.
In machine learning, however, the most critical ingredient of all--the training data--comes 
directly from the outside world.
For a system trained on user data, an attacker can inject malicious data 
simply by creating a user account. Such \emph{data poisoning} attacks require 
us to re-think what it means for a system to be secure.

The focus of the present work is on data poisoning attacks against classification algorithms,
first studied by \citet{biggio2012poisoning} and later by a number of others 
\citep{xiao2012adversarial,xiao2015contamination,newell2014practicality,
mei2015teaching,burkard2017analysis,koh2017understanding}.
This body of work has demonstrated data poisoning attacks that can degrade classifier
accuracy, sometimes dramatically. Moreover, while some defenses have been proposed against specific
attacks \citep{laishram2016curie}, few have been stress-tested against a determined attacker.

Are there defenses that are robust to a large class of data poisoning attacks? 
At development time, one could take a clean dataset and test a 
defense against a number of poisoning strategies on that dataset.
However, because of the near-limitless space of possible attacks, 
it is impossible to conclude from empirical success alone that a defense 
that works against a known set of attacks will not fail against 
a new attack.

In this paper, we address this difficulty by presenting a framework for studying 
the entire space of attacks against a given defense. 
Our framework applies to defenders that (i) remove outliers residing 
outside a feasible set, then (ii) minimize a margin-based loss on the remaining data.
For such defenders, we can generate approximate upper bounds on the efficacy of 
{any} data poisoning attack, which hold modulo two assumptions---that the empirical train and test 
distribution are close together, and that the outlier removal does not significantly 
change the distribution of the clean (non-poisoned) data; these assumptions are detailed more formally in 
Section~\ref{sec:attack-defense}. We then establish a duality result for 
our upper bound, and use this to generate a candidate attack that nearly matches the bound. 
Both the upper bound and attack are generated via an efficient online learning algorithm.

We consider two different instantiations of our framework: first, where the outlier detector 
is trained independently and cannot be affected by the poisoned data, 
and second, where the data poisoning can attack the outlier detector as well.
In both cases we analyze binary SVMs, although our framework applies in the 
multi-class case as well.

In the first setting, we apply our framework to an ``oracle'' defense 
that knows the true class centroids and removes points that are 
far away from the centroid of the corresponding class.
While previous work showed successful attacks on 
the \mnist\ \citep{biggio2012poisoning} 
and Dogfish \citep{koh2017understanding} image datasets in the absence of any defenses, 
we show (Section~\ref{sec:data-independent}) that no attack can substantially increase 
test error against this oracle---the $0/1$-error of an SVM on either dataset is at most 
$4\%$ against any of the attacks we consider, even after adding $30\%$ poisoned 
data.\footnote{We note \citeauthor{koh2017understanding}'s attack on Dogfish targets specific test images 
rather than overall test error.} 
Moreover, we provide certified upper bounds of 
$7\%$ and $10\%$, respectively, on the two datasets.
%\pw{I think people will read this as saying we can defend against the targeted attack on Dogfish. Maybe we want to clarify the type of attack earlier.}
On the other hand, on the IMDB sentiment corpus \citep{maas2011imdb} 
our attack increases classification test error from $12\%$ to $23\%$ with
only $3\%$ poisoned data, showing that defensibility is very dataset-dependent:
the high dimensionality and abundance of irrelevant features in the IMDB corpus 
give the attacker more room to construct attacks that evade outlier removal.

For the second setting, 
we consider a more realistic defender that uses the empirical (poisoned) centroids.
For small amounts of poisoned data ($\leq 5\%$) we can still certify the resilience of 
\mnist{} and Dogfish (Section~\ref{sec:data-dependent}).
However, with more ($30\%$) poisoned data, the attacker can subvert the outlier removal to obtain 
stronger attacks, increasing test error on \mnist{} to $40\%$---much higher than the 
upper bound of $7\%$ for the oracle defense.
In other words, defenses that rely on the (potentially poisoned) data can be much weaker than 
their data-independent counterparts, underscoring the need for outlier removal mechanisms
that are themselves robust to attack.

\section{Problem Setting}
\label{sec:model}
% go over attack

% set up task: loss function
Consider a prediction task from
an input $x \in \sX$ (e.g., $\R^d$) to an output $y \in \sY$; in our case we will take 
$\sY = \{-1,+1\}$ (binary classification) although most of our analysis holds for arbitrary $\sY$.
Let $\ell$ be a non-negative convex loss function: e.g., for linear classification 
with the hinge loss,
$\ell(\theta; x,y) = \max(0, 1 - y\langle \theta, x \rangle)$
for a model $\theta \in \Theta \subseteq \R^d$ and data point $(x,y)$.
Given a true data-generating distribution $p^*$ over $\sX \times \sY$,
define the test loss as $\bL(\theta) = \bE_{(x,y) \sim p^*}[\ell(\theta; x,y)]$.

% Setting
We consider the \emph{causative attack} model \citep{barreno2010security}, 
which consists of a game between two players:
%In this game, there are two players: 
the \emph{defender} (who seeks to learn a model $\theta$),
and the \emph{attacker} (who wants the learner to learn a bad model).
%The capabilities of the attacker and defender are as follows:
The game proceeds as follows:
\begin{itemize}[itemsep=2pt,topsep=0pt,parsep=0pt,partopsep=0pt,leftmargin=18pt]
\item $n$ data points are drawn from $p^*$ to produce a clean training dataset $\sDc$.
\item The attacker adaptively chooses a ``poisoned'' dataset $\sDp$ of $\epsilon n$ poisoned
  points, where $\epsilon \in [0, 1]$ parametrizes the attacker's resources. %controls the resources of the attacker.
\item The defender trains on the full dataset $\sDc \cup \sDp$ to produce a model $\thetaNom$,
      and incurs test loss $\bL(\thetaNom)$.
\end{itemize}
The defender's goal is to minimize the quantity $\bL(\thetaNom)$ while the attacker's goal is to maximize it.
%if $\epsilon$ is very
%small then the attacker only has the ability to insert a small number of points into the
%dataset, while if $\epsilon$ is large the attacker controls a large fraction of the total data.
% say this corresponds to the "causative attack" model
% of Barreno (and explain why)

% attacker: full knowledge of defender's system
\paragraph{Remarks.}
We assume the attacker has full knowledge of the defender's algorithm
and of the clean training data $\sDc$. While this may seem generous to the attacker, it is widely
considered poor practice to rely on secrecy for security \citep{kerckhoffs1883security,biggio2014security};
moreover, a determined attacker can often reverse-engineer necessary system details \citep{tramer2016stealing}.

% brief mention of why this is natural (with some real-world examples)
The causative attack model allows the attacker to add points but not modify existing ones.  Indeed, systems constantly collect new data (e.g., product reviews, user
feedback on social media, or insurance claims), whereas modification of existing data would require first compromising
the system.
% PL: slimmed this down due to space
%sets that are collected online as the system operates, such as product reviews, spam
%detection, sentiment analysis, or user feedback in social networks. In such cases,
%it is easy for an adversary to include new data into the dataset (e.g. by creating
%one or more user accounts).
%On the other hand, it would be difficult to modify already
%existing data without first compromising the system in some other way. While this model already
%encompasses a number of real-world settings, we think it will become even more important
%in the future, as safety-critical systems such as self-driving cars also come to rely
%on user-provided data.
%\todo{provide relevant citations?}

% why maximizing this loss (rather than some other metric)
Attacks that attempt to increase the overall test loss $\bL(\thetaNom)$,
known as \emph{indiscriminate availability} attacks \citep{barreno2010security},
can be thought of as a denial-of-service attack.
%, because it damages the performance of the model in a broad, untargeted sense.
This is in contrast to targeted attacks on individual examples or sub-populations \citep[e.g.,][]{burkard2017analysis}.
Both have serious security implications,
but we focus on denial-of-service attacks, 
as they compromise the model in a broad sense 
and interfere with fundamental statistical properties of learning algorithms.
%\todo{say something sharper here}

%This is only one possible goal of an attacker, and one could consider other goals as well
%(for instance, a spammer might only care about a specific message being classified as non-spam,
%rather than hurting the performance of the system overall).
%While we think these other settings are worth studying, we focused on denial-of-service
%because it leads to a direct conflict between attacker and defender goals, and because a
%successful attack would mean that the system was harmed in a serious and fundamental way.
%rather than exploiting a statistical edge case.

%%%%%%%%%%%%%%%%%%%%%%%%%%%%%%
\subsection{Data Sanitization Defenses}
\label{sec:defense-simple}

% give example attack and defense strategies
A defender who trains na\"ively on the full (clean + poisoned)
data $\sDcp$ is doomed to failure,
as even a single poisoned point can in some cases arbitrarily change the model
\citep{liu2016teaching,park2017resilient}. 
%\pw{This is not true for robust losses (which are defined 
%as having bounded influence functions, so by definition it is not true). Actually, maybe we can say something 
%about how loss bounded by M = at most M leverage? Might motivate studying robust losses.}
%
In this paper, we consider \emph{data sanitization} defenses \citep{cretu2008casting},
which examine the full dataset and try to remove the poisoned points,
for example by deleting outliers. % \citep{laishram2016curie}.
Formally, the defender constructs a \emph{feasible set} $\sF \subseteq \sX \times \sY$
and trains only on points in $\sF$:
\begin{align}
\label{eq:defender-model}
  \thetaNom \eqdef \argmin_{\theta \in \Theta} L(\theta; (\sDcp) \cap \sF), \quad\text{where }
  L(\theta; S) \eqdef \sum_{(x,y) \in S} \ell(\theta; x, y).
\end{align}
Given such a defense $\sF$, we would like to upper bound the
worst possible test loss over \emph{any attacker} (choice of $\sDp$)---in
symbols, $\max_{\sDp} \bL(\thetaNom)$.  Such a bound would certify
that the defender incurs at most some loss no matter what the attacker does. 
We consider two classes of defenses:
\begin{itemize}[itemsep=2pt,topsep=0pt,parsep=0pt,partopsep=0pt,leftmargin=18pt]
  \item \emph{Fixed} defenses, where $\sF$ does not depend on $\sDp$.
    One example for text classification is letting $\sF$ be documents that
    contain only licensed words \citep{newell2014practicality}.
    Other examples are \emph{oracle} defenders that depend on the true distribution $p^*$.
    While such defenders are not implementable in practice,
    they provide bounds: if even an oracle can be attacked,
    then we should be worried. %\todo{maybe make this more precise? maybe cut line?}
  \item \emph{Data-dependent} defenses, where $\sF$ depends on $\sDcp$.
    These defenders try to estimate $p^*$ from $\sDcp$ and thus are implementable in practice.
    However, they open up a new line of attack wherein the attacker
    chooses the poisoned data $\sDp$ to change the feasible set $\sF$. %\pw{`attackable' defenses?}
\end{itemize}

\begin{figure}[t]
\centering
\includegraphics[width=0.45\textwidth]{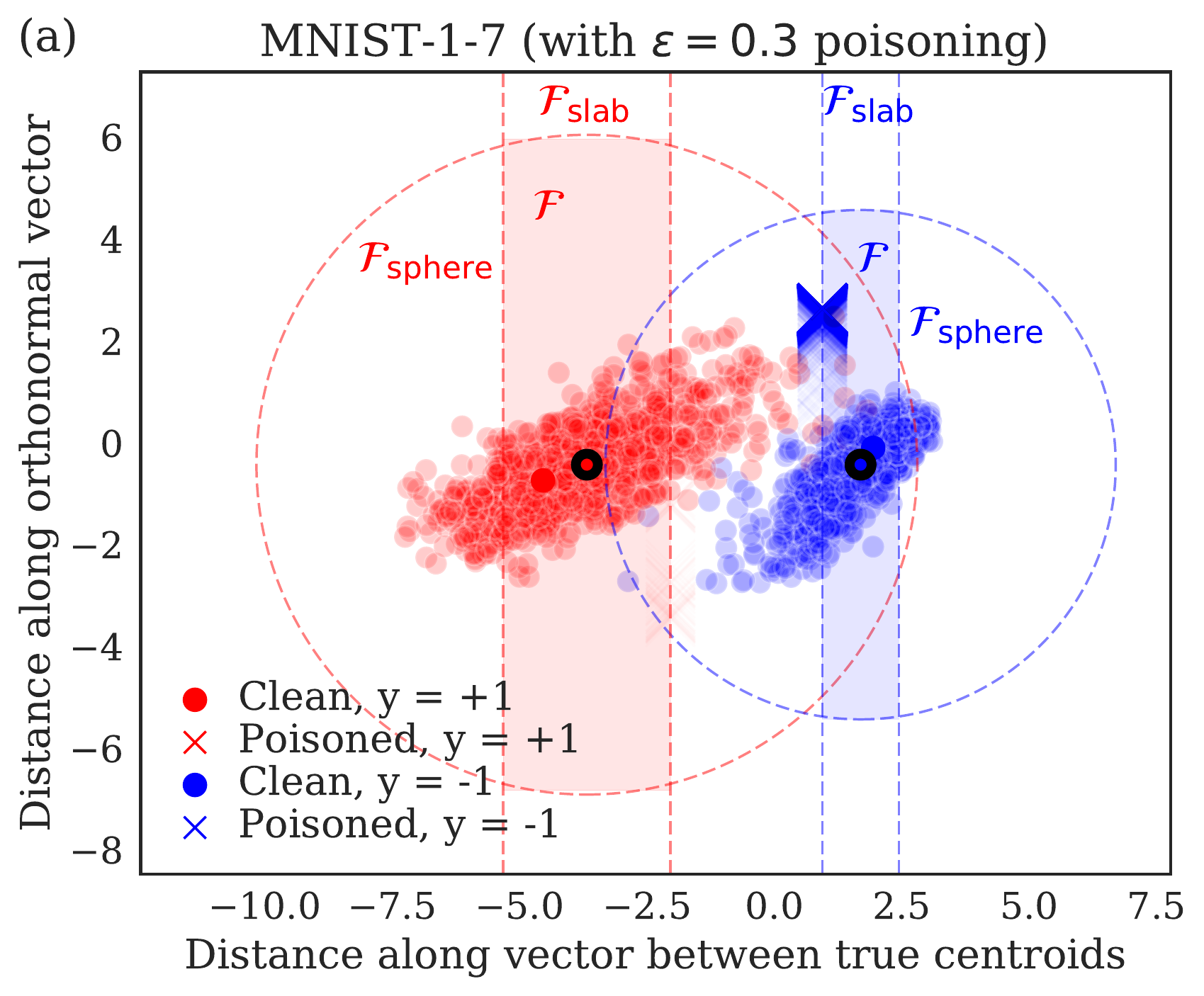}
\includegraphics[width=0.45\textwidth]{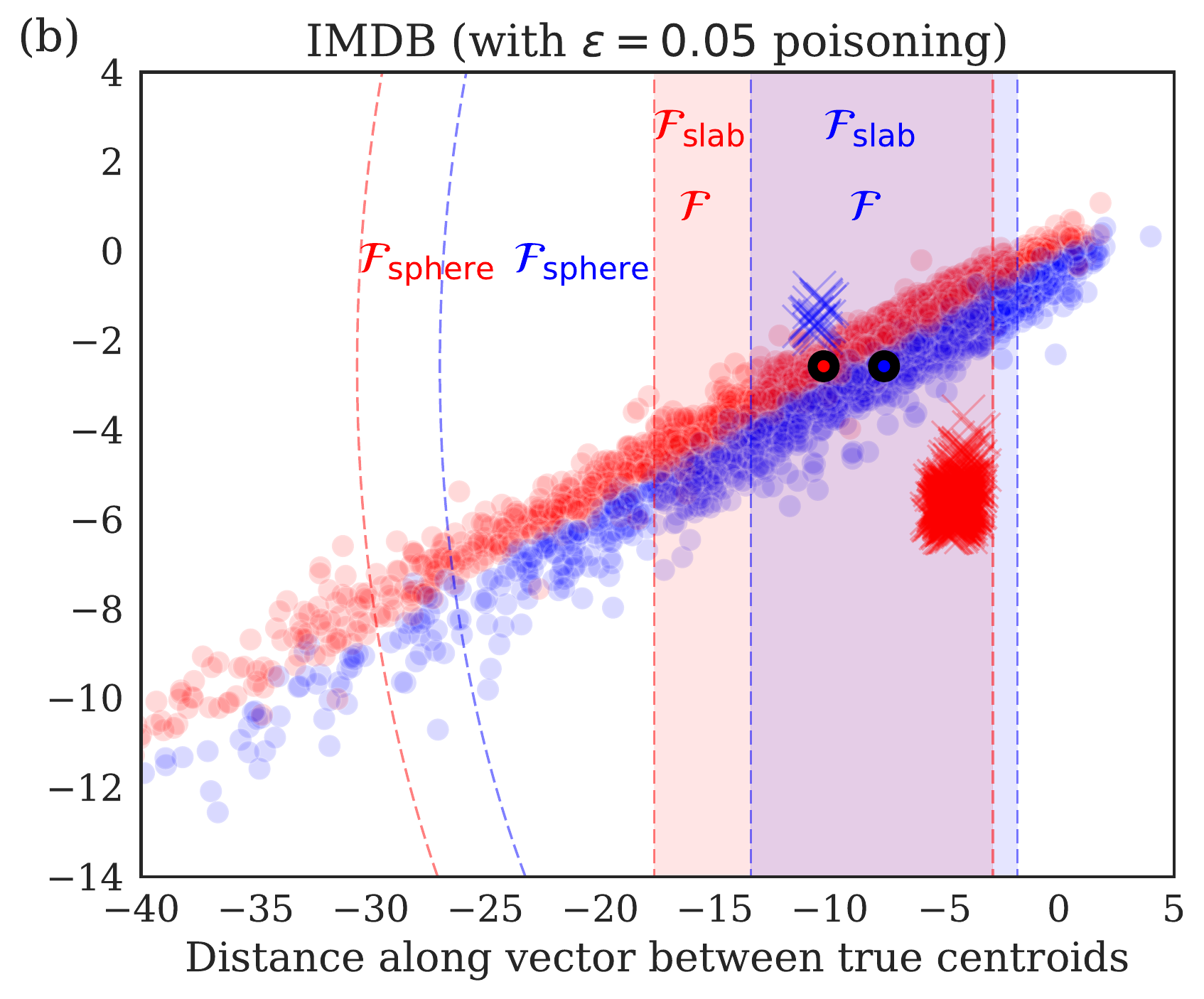}
\vskip -0.03in
  \caption{Different datasets possess very different levels of vulnerability to attack.
  Here, we visualize the effect of the sphere and slab oracle defenses, with thresholds chosen to match the 70th percentile of the clean data.
  We mark with an X our attacks for the respective values of $\epsilon$.
  {\bf(a)} For the \mnist{} dataset, the classes are well-separated and no attack can get past the defense.
  Note that our attack chooses to put all of its weight on the negative class here, although this need not be true in general.
  {\bf(b)} For the IMDB dataset, the class centroids are not well-separated and it is easy to attack the classifier.
  See Section~\ref{sec:data-independent} for more details about the experiments.}
\label{fig:mnist-sphere-slab}
\label{fig:attack-visualization}
\end{figure}

\paragraph{Example defenses for binary classification.} Let
$\mu_+ \eqdef \E[x \mid y = +1]$ and $\mu_- \eqdef \E[x \mid y = -1]$ be the centroids of the
positive and negative classes.
A natural defense strategy is to remove
points that are too far away from the corresponding centroid.
We consider two ways of doing this:
the \emph{sphere defense}, which removes points outside a spherical radius, and
the \emph{slab defense}, which first projects points onto the line between the centroids
and then discards points that are too far on this line:
\begin{align}
\label{eq:defenses}
  \sFsphere \eqdef \{ (x,y) : \|x - \mu_y\|_2 \le r_y \}, \quad
  \sFslab \eqdef \{ (x,y) : |\langle x - \mu_y, \mu_y - \mu_{-y} \rangle| \leq s_y \}.
\end{align}
Here $r_y,s_y$ are thresholds
(e.g. chosen so that 30\% of the data is removed).
Note that both defenses are oracles ($\mu_y$ depends on $p^*$);
in \refsec{slab}, we consider versions that estimate $\mu$ from $\sDcp$.
%\todo{Is 30\% a weird number to throw out here? Seems quite large.}

% introduce sphere + slab constraints
%The first defense, which we call the {\bf sphere} defense, picks thresholds $r_+$ and $r_-$,
%and removes a point $(x,+1)$ if $\|x - \mu_+\|_2 > r_+$, and a point
%$(x,-1)$ if $\|x - \mu_-\|_2 > r_-$. The thresholds $r_+$ and $r_-$ are typically picked to
%remove some percentage, e.g. $10\%$, of the data points in each class. We call this the
%sphere defense because it essentially forces the adversary to place points in a sphere of a
%given radius around the class center $\mu$.

%The second defense, which we call the {\bf slab} defense, projects all points onto the line
%between the two centroids, and discards points that are too far along this line.
%Specifically, it removes a point $(x,+1)$ if $|\langle x - \mu_+, \mu_+ - \mu_- \rangle| > s_+$,
%and analogously for a point $(x,-1)$.
%The motivation is that while Euclidean distance
%may become a poor metric for similarity in high dimensions, the line between the class means
%is likely to be an important direction for many datasets, so protecting this direction from
%attack is useful. We call this the slab defense because the set of feasible points that are not
%removed by the defense forms a slab (see Figure~\ref{fig:mnist-sphere-slab}).

\reffig{mnist-sphere-slab} depicts both defenses on the \mnist{} and IMDB datasets.
% for classifying $1$s versus $7$s,
%assuming that $\mu_+$ and $\mu_-$ are the
%true \pl{on tests data?} (un-poisoned) class centroids, 
%with $r$ and $s$ chosen to match the 70th percentile of the data.
Intuitively, the constraints on \mnist{} make it difficult for an attacker, whereas IMDB 
looks far more attackable. In the next section, we will see how to make these intuitions concrete.

\section{Attack, Defense, and Duality}
% for Data Independent Defenses}
\label{sec:duality}
\label{sec:attack-defense}

%% make point about equilibrium
%In this section, we consider data-independent defenses.
%We wish to certify, for a given defense $\sF$, an upper bound on how
%much damage any attacker could do, as well as provide an actual attack,
%which provides a lower bound.

Recall that we are interested in the worst-case test loss $\max_{\sDp} \bL(\thetaNom)$.
To make progress, we consider three approximations.
First, (i) we pass from the test loss to the training loss on the clean data, 
and (ii) we consider the training loss on the full (clean + poisoned) data,
which upper bounds the loss on the clean data due to non-negativity of the loss.
For any model $\theta$, we then have:
\begin{align}
  \bL(\theta) \stackrel{\text{(i)}}{\approx} \frac{1}{n} L(\theta; \sDc) \stackrel{\text{(ii)}}{\le} \frac{1}{n} L(\theta; \sDcp).
\end{align}
The approximation (i) could potentially be invalid due to overfitting; however, if we 
regularize the model appropriately then we can show that train and test are 
close by standard concentration arguments 
(see Appendix~\ref{sec:overfitting-defense} for details).
Note that (ii) is always a valid upper bound, and will be relatively tight as long 
as the model ends up fitting the poisoned data well.

For our final approximation, we (iii) have the defender train on 
$\sDc \cup (\sDp \cap \sF)$ (i.e., it uses the entire clean data set $\sDc$ 
rather than just the inliers $\sDc \cap \sF$). This should not 
have a large effect as long as the defense is not too aggressive (i.e., as long as 
$\sF$ is not so small that it would remove important points from the clean 
data $\sDc$). We denote the resulting model as $\thetaTil$ to distinguish it 
from $\thetaNom$. 

Putting it all together, the worst-case test loss from any attack $\sDp$ with $\epsilon n$
elements is approximately upper bounded as follows:
\begin{align}
\label{eq:min-max-0}
  \max_{\sDp} \bL(\thetaNom) \stackrel{\text{(i)}}{\approx}
  \max_{\sDp} \frac{1}{n} L(\thetaNom; \sDc) \stackrel{\text{(ii)}}{\le} &
  \max_{\sDp} \frac{1}{n} L(\thetaNom; \sDc \cup (\sDp \cap \sF)) 
\nonumber \\[-1ex]
  \stackrel{\text{(iii)}}{\approx} &
\nonumber  \max_{\sDp} \frac{1}{n} L(\thetaTil; \sDc \cup (\sDp \cap \sF)) \\[-1ex]
  = & \max_{\sDp \subseteq \sF} \min_{\theta \in \Theta} \frac{1}{n} L(\theta; \sDcp) 
\eqdef \bM.
\end{align}
Here the final step is because $\thetaTil$ is chosen to minimize $L(\theta; \sDc \cup (\sDp \cap \sF))$.
The \emph{minimax loss} $\bM$ defined in \eqref{eq:min-max-0} 
is the central quantity that we will focus on in the sequel;
it has duality properties that will yield insight into the nature of the optimal attack.
Intuitively, the attacker that achieves $\bM$ is trying to maximize the loss on the full dataset by adding
poisoned points from the feasible set $\sF$.

The approximations (i) and (iii) define the assumptions we need for our certificates to hold; 
as long as both approximations are valid, $\bM$ will give an approximate upper bound on 
the worst-case test loss.

%%%%%%%%%%%%%%%%%%%%%%%%%%%%%%%%%%%%%%%%%%%%%%%%%%%%%%%%%%%%
%\subsection{Optimal Attacks and Certified Defenses via Online Learning}
\subsection{Fixed Defenses: Computing the Minimax Loss via Online Learning}
\label{sec:data-independent-framework}
% new subsection: solving the min-max problem, via online learning

We now focus on computing the minimax loss $\bM$ in \eqref{eq:min-max-0} when
$\sF$ is not affected by $\sD_p$ (fixed defenses).
%, which provides an approximate upper bound on the worst-case test loss. 
In the process of computing $\bM$,
we will also produce candidate attacks.
Our algorithm is based on no-regret online learning,
which models a game between a learner and nature
and thus is a natural fit to our data poisoning setting.
For simplicity of exposition we assume $\Theta$ is an $\ell_2$-ball of radius $\rho$.
%%We assume for now that $\sF$ is fixed (independent of $\sD_p$). 
%\todo{maybe save a line here?}

\newcommand{\upper}{U^*}
\begin{algorithm}[t!]
\caption{Online learning algorithm for generating an upper bound and candidate attack.}
\label{alg:main}
\begin{algorithmic}
\State {\bf Input:} clean data $\sDc$ of size $n$, feasible set $\sF$, 
       radius $\rho$, %\pl{can we generalize to arbitrary parameter $\Theta$} \js{not without changing algo}, 
       poisoned fraction $\epsilon$, step size $\eta$. %\todo{gap between radius $\rho$ here vs. arbitrary set $\Theta$}
%\State Output: $\upper$, candidate attack $\{(x^{(t)}, y^{(t)})\}_{t=1}^{\epsilon n}$.
%\State Hyperparameter: initial step size $\eta$.
\State Initialize $z^{(0)} \gets 0$, $\lambda^{(0)} \gets \frac{1}{\eta}$, $\theta^{(0)} \gets 0$, $U^* \gets \infty$.
\For{$t = 1, \ldots, \epsilon n$} %\pw{this doesn't converge for small $\epsilon$}
  \State Compute $(x^{(t)}, y^{(t)}) = \argmax_{(x,y) \in \sF} \ell(\theta^{(t-1)}; x,y)$. % ($\dagger$). \todo{this is ugly; also, fix $\dagger$}
  \State $\upper \gets \min\big(\upper, \frac{1}{n} L(\theta^{(t-1)}; \sDc) + \epsilon \ell(\theta^{(t-1)}; x^{(t)}, y^{(t)})\big)$.
  \State $g^{(t)} \gets \frac{1}{n} \nabla L(\theta^{(t-1)}; \sDc) + \epsilon \nabla \ell(\theta^{(t-1)}; x^{(t)}, y^{(t)})$.
  \State Update: $\ \ z^{(t)} \gets z^{(t-1)} - g^{(t)}, \quad \lambda^{(t)} \gets \max(\lambda^{(t-1)}, \frac{\|z^{(t)}\|_2}{\rho}), \quad \theta^{(t)} \gets \frac{z^{(t)}}{\lambda^{(t)}}$.
\EndFor
\State {\bf Output:} upper bound $\upper$ and candidate attack $\sDp = \{(x^{(t)}, y^{(t)})\}_{t=1}^{\epsilon n}$.
\end{algorithmic}
\end{algorithm}

Our algorithm, shown in Algorithm~\ref{alg:main}, is very simple:
in each iteration,
it alternates between finding the worst attack point $(x^{(t)}, y^{(t)})$ with respect to the
current model $\theta^{(t-1)}$ and updating the model in the direction of the 
attack point, producing $\theta^{(t)}$.
The attack $\sDp$ is the set of points thus found.

To derive the algorithm, we simply swap min and max in \eqref{eq:min-max-0}
to get an upper bound on $\bM$, after which the optimal attack set $\sDp \subseteq \sF$ for a \emph{fixed} $\theta$ is realized
by a single point $(x,y) \in \sF$:
\begin{align}
\label{eq:M-to-U}
  \bM
  %&\stackrel{\eqref{eq:min-max-0}}{=} \max_{\sDp \subseteq \sF} \min_{\theta \in \Theta} L(\theta; \sDcp)
      \le \min_{\theta \in \Theta} \max_{\sDp \subseteq \sF} \frac{1}{n} L(\theta; \sDcp)
      %= \min_{\theta \in \Theta} \underbrace{L(\theta; \sDc) + \epsilon n \max_{(x,y) \in \sF} \ell(\theta; x, y)}_{\eqdef U(\theta)}.
      = \min_{\theta \in \Theta} U(\theta), \text{ where } U(\theta) \eqdef \frac{1}{n} L(\theta; \sDc) + \epsilon \max_{(x,y) \in \sF} \ell(\theta; x, y).
\end{align}
% Get upper/lower bounds anyway
Note that $U(\theta)$ upper bounds $\bM$ for any model $\theta$.
Algorithm~\ref{alg:main} follows the natural strategy of minimizing $U(\theta)$
to iteratively tighten this upper bound.
In the process, the iterates $\{(x^{(t)}, y^{(t)})\}$
form a candidate attack $\sDp$
whose induced loss $\frac{1}{n} L(\thetaTil; \sDcp)$ is a lower bound on $\bM$. 
We can monitor the duality gap between lower and upper bounds on $\bM$
to ascertain the quality of the bounds.

% Convex case: tight
Moreover, since the loss $\ell$ is convex in $\theta$,
$U(\theta)$ is convex in $\theta$ (regardless of the structure of $\sF$, which could even be discrete).
In this case, if we minimize $U(\theta)$ using any online learning algorithm with sublinear regret,
the duality gap vanishes for large datasets. 
%\pw{This is the first time we mention needing convexity; should we mention
%this earlier (when we say we can get attacks that nearly match the optimal)? Also would be nice to summarize
%at some point what conditions we require for what guarantees.} \js{I think no due to space reasons}
In particular (proof in Appendix~\ref{sec:regret-proof}):
\newcommand{\Regret}{\operatorname{Regret}}
\begin{proposition}
\label{prop:regret}
Assume the loss $\ell$ is convex.
Suppose that an online learning algorithm (e.g., Algorithm~\ref{alg:main}) is used to minimize $U(\theta)$,
and that the parameters $(x^{(t)}, y^{(t)})$ maximize the loss $\ell(\theta^{(t-1)}; x,y)$ for the iterates
$\theta^{(t-1)}$ of the online learning algorithm.
Let $U^* = \min_{t=1}^{\epsilon n} U(\theta^{(t)})$.
Also suppose that the learning algorithm has regret $\Regret(T)$ after $T$ time steps.
Then, for the attack $\sDp = \{(x^{(t)}, y^{(t)})\}_{t=1}^{\epsilon n}$, 
the corresponding parameter $\thetaTil$ satisfies:
%\js{probably want to clean up this statement}
%\pw{Mention proof in appendix?}
\begin{align}
  \frac{1}{n} L(\thetaTil; \sDcp) \le \bM \le U^* %U(\theta^\dagger)
   \quad\text{and}\quad
  %U(\theta^\dagger) 
  U^* - \frac{1}{n} L(\thetaTil; \sDcp) \le \frac{\Regret(\epsilon n)}{\epsilon n}.
\end{align}
\end{proposition}
\vskip -0.1in
Hence, any algorithm whose average regret $\frac{\Regret(\epsilon n)}{\epsilon n}$ 
is small will have a nearly optimal candidate attack $\sDp$.
%Algorithm~\ref{alg:main} could employ any no-regret learning algorithm to
%compute the iterates $\theta^{(t)}$; 
There are many algorithms that have this property
%, usually attaining average regret $\oo(1/\sqrt{\epsilon n})$ or better 
\citep{shalev2011online}; 
the particular algorithm depicted in Algorithm~\ref{alg:main} is a variant of 
regularized dual averaging \citep{xiao2010rda}.
%, which is an alternative to gradient descent that exploits 
%the norm constraint on $\Theta$ to converge faster.
%\pw{First mention of norm constraint, other than in alg 1 (where it appears 
%without warning.}
%
In summary, we have a simple learning algorithm that computes an upper bound
on the minimax loss along with a candidate attack (which provides a lower bound).
Of course, the minimax loss $\bM$ is only an approximation to the true worst-case test loss
(via \eqref{eq:min-max-0}).
We examine the tightness of this approximation empirically in Section~\ref{sec:data-independent}.

%%%%%%%%%%%%%%%%%%%%%%%%%%%%%%%%%%%%%%%%%%%%%%%%%%%%%%%%%%%%

\subsection{Data-Dependent Defenses: Upper and Lower Bounds}
\label{sec:data-dependent-analysis}

We now turn our attention to data-dependent defenders, where 
the feasible set $\sF$ depends on the data $\sDcp$ (and hence can be influenced 
by the attacker). For example, consider the slab defense (see \eqref{eq:defenses}) that uses the empirical 
(poisoned) mean instead of the true mean:
\begin{equation}
\sFslab(\sDp) \eqdef \{(x,y) : |\langle x - \hat{\mu}_y(\sDp), \hat{\mu}_y(\sDp) - \hat{\mu}_{-y}(\sDp) \rangle| \leq s_y \},
\end{equation}
where $\hat{\mu}_y(\sDp)$ is the empirical mean over $\sDcp$;
the notation $\sF(\sDp)$ tracks the dependence of the feasible set on $\sDp$.
%($s_y$ also depends on $\sDp$, but we ignore this).
%
Similarly to Section~\ref{sec:data-independent-framework}, we analyze the
minimax loss $\bM$, which we can bound as in \eqref{eq:M-to-U}:
%upper bound 
%\eqref{eq:M-to-U} on the minimax loss $\bM$:
%\begin{align}
  $\bM \le \min_{\theta \in \Theta} \max_{\sDp \subseteq \sF(\sDp)} \frac{1}{n} L(\theta; \sDcp)$.
%\end{align}

However, unlike in \eqref{eq:M-to-U}, it is no longer the case that the 
optimal $\sDp$ places all points at a single location, due to the dependence of $\sF$ on $\sDp$; 
we must jointly maximize over the full set $\sDp$.
To improve tractability, we take a continuous relaxation: we think of $\sDp$ as a 
probability distribution with mass $\frac{1}{\epsilon n}$ on each point in $\sDp$, 
and relax this to allow any probability distribution $\pip$. The constraint 
then becomes $\support(\pip) \subseteq \sF(\sDp)$ (where $\support$ denotes the support), 
and the analogue to \eqref{eq:M-to-U} is
\begin{align}
\label{eq:M-to-U-2}
  \bM \le \min_{\theta \in \Theta} \Up(\theta), \text{ where } \Up(\theta) \eqdef \frac{1}{n} L(\theta; \sDc) + \epsilon \max_{\support(\pip) \subseteq \sF(\pip)} \bE_{\pip}[\ell(\theta; x, y)].
\end{align}
This suggests again employing Algorithm~\ref{alg:main} to 
minimize $\Up(\theta)$. Indeed, this is what we shall do, but there are a few caveats:
\begin{itemize}[itemsep=2pt,topsep=0pt,parsep=0pt,partopsep=0pt,leftmargin=18pt]
\item The maximization problem in the definition of $\Up(\theta)$ is in general 
  quite difficult. We will, however, solve a specific instance in 
      Section~\ref{sec:data-dependent} based on the sphere/slab defense described in 
      Section~\ref{sec:defense-simple}.
      %Nevertheless, we will see that when $\ell(\theta; x,y)$ is 
      %the hinge loss, there is a tractable algorithm based on semidefinite programming.
\item The constraint set for $\pip$ is non-convex, so duality (Proposition~\ref{prop:regret}) 
      no longer holds. In particular, the average of two feasible $\pip$ might not itself 
      be feasible. %\todo{example of why it doesn't hold?}
\end{itemize}
To partially address the second issue, 
we will run Algorithm~\ref{alg:main}, at each iteration obtaining a distribution 
$\pip^{(t)}$ and upper bound $\Up(\theta^{(t)})$. 
Then, for \emph{each} $\pip^{(t)}$ we will generate a candidate attack 
by sampling $\epsilon n$ points from $\pip^{(t)}$, and take the best resulting attack.
In Section~\ref{sec:data-independent} we will see that despite a lack of rigorous theoretical 
guarantees, this often leads to good upper bounds and attacks in practice.
%\todo{needs a bit of work}
%It is tempting to 
%average together all of the distributions $\pip^{(1)}$, $\ldots$, $\pip^{(T)}$ before 
%sampling, but because the constraint $\support(\pip) \subseteq \sF(\pip)$ is non-convex, 
%this might lead to an infeasible attack.
%\todo{explain this better}

{
\setlength{\belowcaptionskip}{-6pt}
\begin{figure}
\begin{center}
\includegraphics[width=\textwidth]{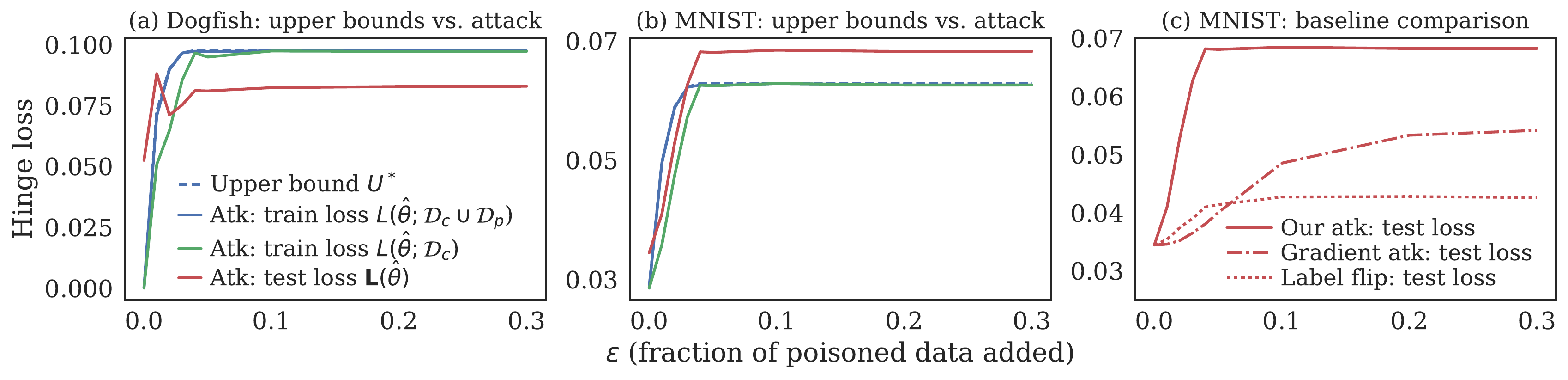}
\end{center}
\vskip -0.09in
\caption{
  On the {\bf (a)} Dogfish and {\bf (b)} \mnist{} datasets, our candidate attack (solid blue) achieves the upper bound (dashed blue)  
	on the worst-case train loss, as guaranteed by Proposition~\ref{prop:regret}. % under the oracle sphere and slab defense. 
	Moreover, this worst-case loss is low; even after adding 30\% poisoned data, the loss stays below 0.1.
	{\bf (c)} The gradient descent (dash-dotted) and label flip (dotted) baseline attacks are suboptimal under this defense, with 
	test loss (red) as well as test error and train loss (not shown) all significantly worse than our candidate attack.}
\label{fig:results-mnist}
\label{fig:results-dogfish}
\end{figure}
}

%\begin{figure}
%\begin{center}
%\includegraphics[width=0.32\textwidth]{figures/fig1-dogfish.pdf}
%\includegraphics[width=0.32\textwidth]{figures/fig1-mnist_17.pdf}
%%\includegraphics[width=0.32\textwidth]{figures/fig2-mnist_17.pdf}
%\includegraphics[width=0.32\textwidth]{figures/fig3-mnist_17.pdf}
%\end{center}
%\caption{Results on the MNIST dataset (1 vs. 7). (a) Plot of the upper bound on the 
%training loss computed by our algorithm (blue dashed), together with the actual training loss 
%of the resulting attack (blue solid), the training loss on just the clean data 
%(green), and the test loss (purple). (b) Comparison of our attack to two baselines. 
%(c) Comparison to the same baselines on the test set, measuring both hinge and $0/1$-loss.}
%\label{fig:results-mnist}
%\label{fig:results-dogfish}
%\end{figure}

\section{Experiments I: Oracle Defenses}
\label{sec:data-independent}
%
%\subsection{Experiments on Image Data}
\label{sec:image-experiments}
% experiments on image data:
% to illustrate our framework, we run experiments on the dogfish and mnist datasets
An advantage of our framework is that we obtain a tool that can be easily run on new 
datasets and defenses to learn about the robustness of the defense and gain insight 
into potential attacks. 
We first study two image datasets: \mnist{}, 
and the Dogfish dataset used by \citet{koh2017understanding}. For \mnist{}, following
\citet{biggio2012poisoning}, we considered binary classification between the digits
$1$ and $7$; this left us with $n = 13007$ training examples of dimension $784$.
For Dogfish, which is a binary classification task, we used the same Inception-v3 
features as in \citet{koh2017understanding}, so that each of the $n = 1800$ 
training images is represented by a $2048$-dimensional vector.
%\footnote{While
%we attack the neural net representation, previous work \citep{dosovitskiy2016inverting, mahendran2015understanding}
%has studied how to invert neural networks, i.e., reconstruct the input from the representation.} 
For this and subsequent experiments, our loss $\ell$ is the hinge loss (i.e., we 
train an SVM).  

% for our learning algorithm, run our algorithm with RDA as optimization
% consider sphere + slab defenses, use slightly more sophisticated version that
We consider the combined oracle slab and sphere defense from 
Section~\ref{sec:defense-simple}: $\sF = \sFslab \cap \sFsphere$. 
To run Algorithm~\ref{alg:main}, we need to maximize the loss over 
$(x,y) \in \sF$.
Note that maximizing the hinge loss $\ell(\theta; x, y)$ is equivalent to minimizing $y \langle \theta, x \rangle$.
Therefore, we can solve the following quadratic program (QP) for each
$y \in \{+1,-1\}$ and take the one with higher loss:
\begin{align}
\label{eq:qp-1}
%\text{maximize}\ & \max(1 - \langle \theta, x \rangle, 0) 
% PL: make this actually a QP
\text{minimize}_{x \in \bR^d} \ \ & y \langle \theta, x \rangle
\quad \text{subject to}\  \|x - \mu_y\|_2^2 \leq r_y^2, \quad |\langle x - \mu_y, \mu_y - \mu_{-y} \rangle| \leq s_y.
\end{align}
%$\langle \theta, x \rangle$, and hence \eqref{eq:qp-1} can be solved efficiently.
%Finding the optimum for $y_p \in \{-1,+1\}$ in turn allows us
%to solve ($\dagger$). In addition, while QPs in high dimensions can be computationally expensive, the optimum
%of \eqref{eq:qp-1} always lies in the span of $\theta$, $\mu_+$, and $\mu_-$, and hence we can actually solve
%a $3$-dimensional QP, which is extremely fast.
The results of Algorithm~\ref{alg:main} are given in 
Figures~\ref{fig:results-mnist}a and \ref{fig:results-dogfish}b;
here and elsewhere, we used a combination of CVXPY \citep{diamond2016cvxpy}, YALMIP \citep{lofberg2004}, SeDuMi \citep{sturm1999guide}, 
and Gurobi \citep{gurobi2016} to solve the optimization. 
We plot the upper bound $U^*$ computed by Algorithm~\ref{alg:main}, as well as
the train and test loss induced by the corresponding attack $\sDp$.
% chooses threshold adaptively based on cross-validation, then solves the constrained SVM
% [SKIP] (importantly, we *DONT* cross-validate SVM regularizer b/c this opens up to overfitting attacks)
%\todo{say what algorithm we ran for the learner}
%The defender operates by first sanitizing data with the sphere and slab defenses (adaptively
%choosing a quantile cutoff in $[0.7,1.0]$ via cross-validation) and then trains an SVM
%on the remaining data. % according to \eqref{eq:svm}.
%
%\input results-dogfish-figure
%
% go over results
%For both datasets, $U^*$ bound closely matches the loss on the full data 
%$\sDcp$ (as guaranteed by Proposition~\ref{prop:regret}). Moreover, 
Except for small $\epsilon$, 
the model $\thetaTil$ fits the poisoned data almost perfectly.
We think this is because all feasible attack points that can get past the 
defense can be easily fit without sacrificing the quality of the rest of the 
model; in particular, the model chooses to fit the attack points as soon as 
$\epsilon$ is large enough that there is incentive to do so.

The upshot is that, in this case, the loss $L(\thetaTil; \sDc)$ on the 
clean data nearly matches its upper bound $L(\thetaTil; \sDcp)$ (which in turn matches $U^*$). 
On both datasets, the certified upper bound 
$U^*$ is small ($<0.1$ with $\epsilon=0.3$), showing that the datasets 
are resilient to attack under the oracle defense.

We also ran the candidate attack from Algorithm~\ref{alg:main} 
as well as two baselines --- gradient descent on the test loss (varying the location 
of points in $\sDp$, as in \citet{biggio2012poisoning} and \citet{mei2015teaching}), 
and a simple baseline
that inserts copies of points from $\sDc$ with the opposite label (subject
to the flipped points lying in $\sF$).
The results are in in Figure~\ref{fig:results-mnist}c. 
Our attack consistently performs strongest; 
label flipping seems to be too weak, while the gradient algorithm seems to get 
stuck in local minima.\footnote{Though \citet{mei2015teaching} state 
that their cost is convex, they communicated to us that this is incorrect.}
Though it is not shown in the figure, we note that 
the maximum test 0-1 error against any attack, for $\epsilon$ up to 0.3, was 4\%, 
confirming the robustness suggested by our certificates.

Finally, we visualize our attack in Figure~\ref{fig:attack-visualization}a.
Interestingly, though the attack was free to place points anywhere, 
most of the attack is tightly concentrated around a single point at the boundary 
of $\sF$.

\begin{figure}
\begin{center}
\includegraphics[width=\textwidth,height=1.7in]{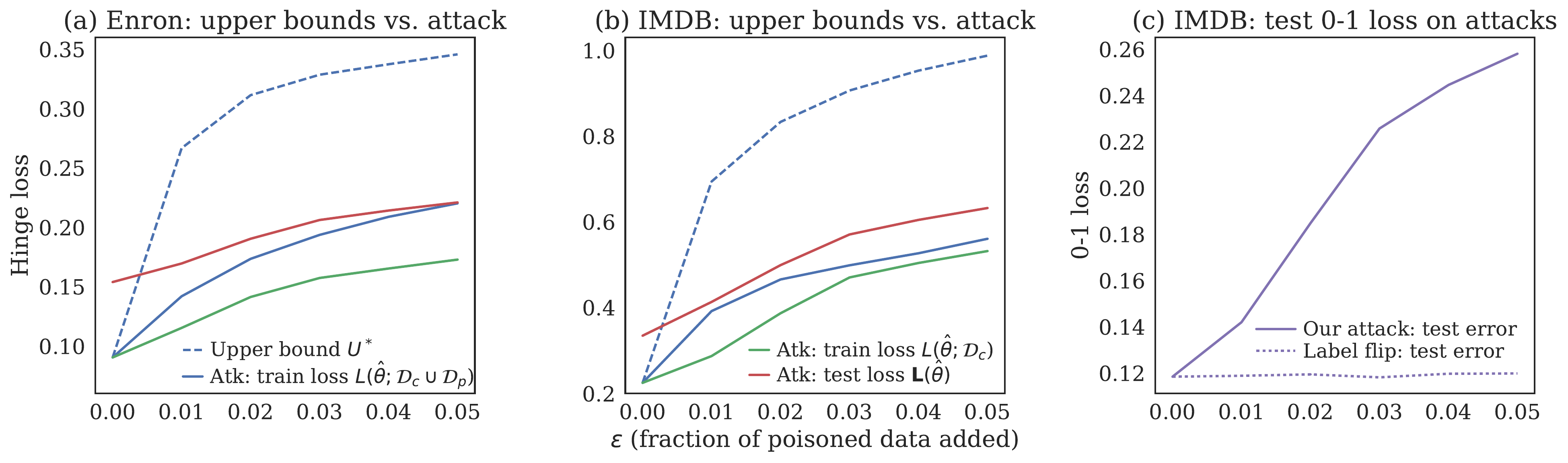}
\end{center}
\vskip -0.09in
  \caption{The {\bf (a)} Enron and {\bf (b)} IMDB text datasets are significantly easier to attack under the oracle sphere and slab defense
  than the image datasets from Figure~\ref{fig:results-mnist}. {\bf (c}) In particular, our attack achieves a large increase in test loss (solid red) and test error (solid purple) with small 
  $\epsilon$ for IMDB. The label flip baseline was unsuccessful as before, and the gradient 
  baseline does not apply to discrete data.
  In (a) and (b), note the large gap between upper and lower bounds, resulting from the
  upper bound relaxation and the IQP/randomized rounding approximations.
  }
\label{fig:results-enron}
\end{figure}

%
%\begin{figure}
%\begin{center}
%\includegraphics[width=0.32\textwidth]{figures/fig4-enron.pdf}
%\includegraphics[width=0.32\textwidth]{figures/fig4-imdb.pdf}
%\includegraphics[width=0.32\textwidth]{figures/fig6-imdb.pdf}
%%\includegraphics[width=0.32\textwidth]{figures/fig5-enron.pdf}
%%\includegraphics[width=0.32\textwidth]{figures/fig6-enron.pdf}
%\end{center}
%  \caption{Results on the Enron spam dataset. \pl{better caption that explains things and punchline}
%  \pl{write 'upper' and 'lower'?}}
%\label{fig:results-enron}
%\end{figure}

% new section: handling input constraints
\subsection{Text Data: Handling Integrity Constraints}
\label{sec:input-constraints}

We next consider attacks on text data. Beyond the 
the sphere and slab constraints, a valid attack on text data must satisfy 
additional \emph{integrity constraints} \citep{newell2014practicality}:
for text, the input $x$ consists of binary indicator features (e.g., presence of the word ``banana'') 
rather than arbitrary reals.\footnote{Note that in the previous section, we ignored such 
integrity constraints for simplicity.}
%Indeed, even in the previous section, an actual attack 
%on the Dogfish dataset would need to reverse-engineer images that lead to a specific 
%neural net representation, an issue which we ignored for simplicity.} \pw{what about mnist non-negativity?}

Algorithm~\ref{alg:main} still applies in this case ---
the only difference is that the QP from Section~\ref{sec:image-experiments} has the added 
constraint $x \in \bZ_{\geq 0}^d$ and hence becomes an integer quadratic program (IQP), 
which can be computationally expensive to solve. 
We can still obtain upper bounds simply by relaxing the integrity constraints; 
the only issue is that the points $x^{(t)}$ in the corresponding attack 
will have continuous values, and hence don't correspond to actual text inputs.
To address this, we use the IQP solver from Gurobi \citep{gurobi2016} 
to find an approximately optimal feasible $x$. This yields a valid candidate attack, but it might 
not be optimal if the solver doesn't find near-optimal solutions.

We ran both the upper bound relaxation and the IQP solver on two text datasets, 
the Enron spam corpus \citep{metsis2006spam} and the IMDB sentiment corpus \citep{maas2011imdb}.
The Enron training set consists of $n = 4137$ e-mails ($30\%$ 
spam and $70\%$ non-spam), with $d = 5166$ distinct words. The IMDB training set 
consists of $n = 25000$ product reviews 
with $d = 89527$ distinct words. We used bag-of-words features, 
which yields test accuracy $97\%$ and $88\%$, respectively, in the absence 
of poisoned data.
%For these experiments we took the poisoning fraction $\epsilon \in [0, 0.05]$. 
IMDB was too large for Gurobi to even approximately solve the IQP, so 
we resorted to a randomized rounding heuristic to convert the continuous relaxation 
to an integer solution. 

Results are given in Figure~\ref{fig:results-enron}; 
there is a relatively large gap between 
the upper bound %(blue dashed) 
and the attack. % due to difficulty in solving the IQP. % (blue solid). 
%This is due to the computational difficulty of solving the IQP, 
%which leads to points that are far from optimal; it may also 
%be due in part to conservatism in the relaxed upper bound.
%
Despite this, the attacks are relatively successful. Most striking 
is the attack on IMDB, which increases test error from $12\%$ to $23\%$ 
for $\epsilon = 0.03$, despite having to pass the oracle defender. 
%More accurate solutions to the IQP would likely 
%lead to even larger increases in test error. 

To understand why the attacks are so much more successful in this 
case, we can consult Figure~\ref{fig:attack-visualization}b. In contrast 
to \mnist{}, for IMDB the defenses place 
few constraints on the attacker. This seems to be a consequence 
of the high dimensionality of IMDB and the large number of irrelevant 
features, which increase the size of $\sF$ without a corresponding 
increase in separation between the classes. %\todo{save line?} \todo{say more about the attack?}

\section{Experiments II: Data-Dependent Defenses}
\label{sec:slab}
\label{sec:slab-attack}
\label{sec:data-dependent}

{
\setlength{\belowcaptionskip}{-6pt}
\begin{figure}
\begin{center}
\includegraphics[width=0.95\textwidth]{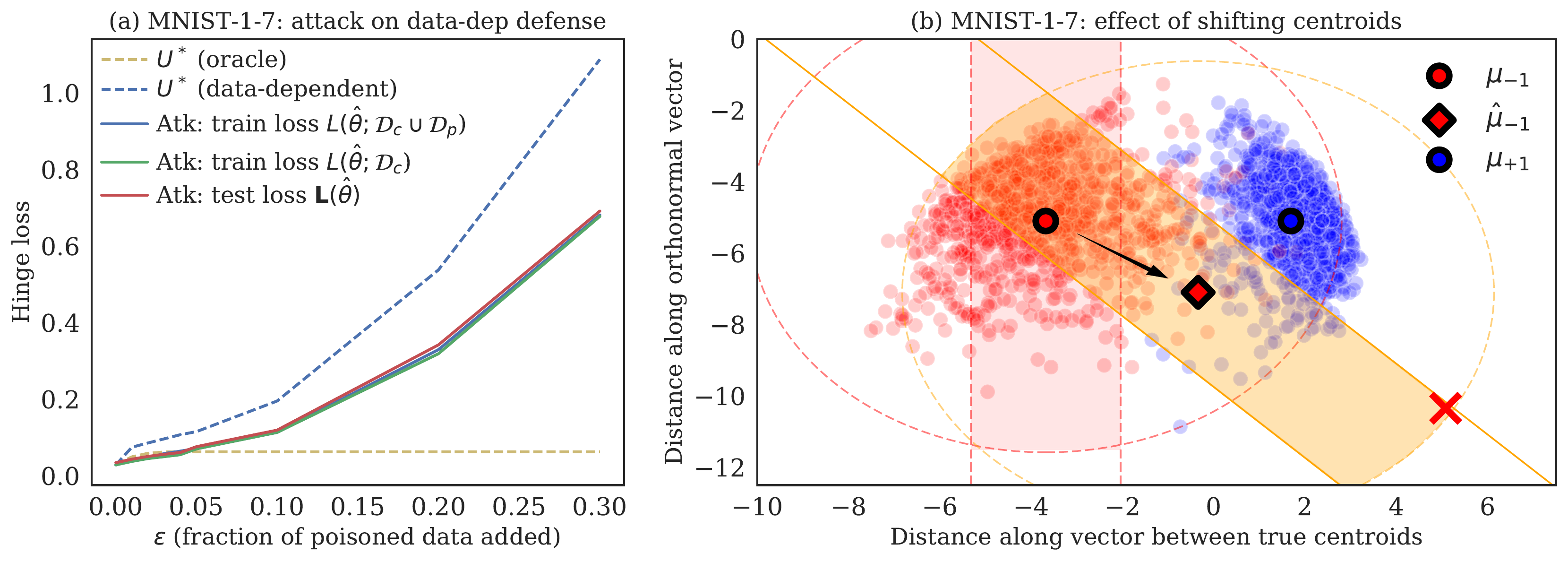}
\end{center}
\vskip -0.06in
\caption{
	The data-dependent sphere and slab defense is significantly weaker than its oracle counterpart,
	allowing \mnist{} and Dogfish to be successfully attacked.
	{\bf (a)} On \mnist{}, our attack achieves a test loss of 0.69 (red) and error of 0.40 (not shown)
	at $\epsilon=0.3$, more than $10\times$ its oracle counterpart (gold). At low $\epsilon \leq 0.05$, the dataset is
	safe, with a max train loss of 0.12. We saw qualitatively similar results on Dogfish.
	{\bf (b)} Data-dependent sanitization can be significantly poisoned by coordinated adversarial data.
	We show here our attack for $\epsilon=0.3$, which places almost all of its attacking mass
	on the red X. This shifts the empirical centroid, rotating the slab constraint (from red to orange)
  and allowing the red X to be placed far on the other side of the blue centroid.}
\label{fig:results-dogfish-slab}
\end{figure}
}

We now revisit the \mnist{} and Dogfish datasets.
Before, we saw that they 
were unattackable provided we had an oracle defender that 
knew the true class means. If we instead consider a data-dependent 
defender that uses the empirical (poisoned) means, 
how much can this change the attackability of these datasets?
In this section, we will see that the answer is quite a lot.

As described in Section~\ref{sec:data-dependent-analysis}, we can still use 
our framework to obtain upper and lower bounds even in this data-dependent case, 
although the bounds won't necessarily match. The main difficulty is in computing 
$\Up(\theta)$, which involves a potentially intractable maximization 
(see \eqref{eq:M-to-U-2}). However, for $2$-class SVMs there is a tractable 
semidefinite programming algorithm; the full details are in Appendix~\ref{sec:sdp}, 
but the rough idea is the following: we can show that the optimal distribution 
$\pip$ in \eqref{eq:M-to-U-2} is supported on at most $4$ points 
(one support vector and one non-support vector in each class). Moreover, for
a fixed $\pip$, the constraints and objective 
depend only on inner products between a small number of points: 
the $4$ attack points, the class means $\mu$ (on the clean data), and the model $\theta$.
Thus, we can solve for the optimal attack locations with a semidefinite 
program on a $7 \times 7$ matrix. Then in an outer loop, we
randomly sample $\pip$ from the probability simplex
and take the one with the highest loss.
Running this algorithm on \mnist{} yields the results in 
Figure~\ref{fig:results-dogfish-slab}a.
On the test set, our $\epsilon=0.3$ attack
leads to a hinge loss of $0.69$ (up from $0.03$) and a 0-1 loss of $0.40$ (up from $0.01$).
Similarly, on Dogfish, our $\epsilon=0.3$ attack gives a hinge loss of 
$0.59$ (up from $0.05$) and a 0-1 loss of $0.22$ (up from $0.01$).

The geometry of the attack is depicted in Figure~\ref{fig:results-dogfish-slab}b. 
By carefully choosing the location of the attack, the attacker can place points that 
lie substantially outside the original (clean) feasible set. This is because 
the poisoned data can substantially change the the direction 
of the slab constraint, while the sphere constraint by itself is not enough to 
effectively filter out attacks. 
There thus appears to be significant danger in employing data-dependent 
defenders---beyond the greater difficulty of analyzing them, they seem to 
actually be more vulnerable to attack.

% test on dogfish, MNIST datasets
% interpret results
% visualize attack

\section{Related Work}

Due to their increased use in security-critical settings such as malware detection, 
there has been an explosion of work on the security of machine learning systems; 
see \citet{barreno2010security}, \citet{biggio2014security}, 
\citet{papernot2016towards}, and \citet{gardiner2016security} for some recent surveys.

Our contribution relates to the long 
line of work on data poisoning attacks;
beyond linear classifiers, others have studied the 
%\citep{biggio2011label,biggio2012poisoning,biggio2014securitysvm,
%xiao2012adversarial,xiao2015contamination,
%newell2014practicality,
%mei2015teaching,laishram2016curie,burkard2017analysis,koh2017understanding,park2017resilient}, 
LASSO \citep{xiao2015lasso}, clustering \citep{biggio2013clustering,biggio2014linkage}, 
PCA \citep{rubinstein2009antidote}, 
topic modeling \citep{mei2015security}, 
collaborative filtering \citep{li2016data}, neural networks 
\citep{yang2017generative}, and other models \citep{mozaffari2015systematic,vuurens2011spam,wang2016combating}. 
%Beyond showing proof-of-concept attacks against various popular 
%machine learning algorithms, 
There have also been a number of 
demonstrated vulnerabilities in deployed systems 
\citep{newsome2006paragraph,laskov2014practical,biggio2014malware}.
We provide formal scaffolding to this line of work by supplying
a tool that can certify defenses against a range of attacks.
% modulo 
%well-defined assumptions.
% as well as automatically 
%generating generating nearly optimal attacks. 
%
%While previous work has considered 
%automatically generated attacks based on gradient descent 
%\citep{biggio2012poisoning,mei2015teaching}, 
%alternating minimization \citep{xiao2012adversarial,xiao2015contamination}, and other methods, 
%these are all non-convex local methods, and can get stuck in local optima as shown in 
%Section~\ref{sec:image-experiments}.

%Beyond traditional machine learning systems, data poisoning has been studied in the 
%context of crowdsourcing 
%as well as analyzed as a Stackelberg game \citep{bruckner2012static,zhou2016modeling}. 
%\todo{expand or shorten?}
%One tool used in such approaches is \emph{bilevel programming} \citep{bard1999}, 
%in which one maximizes an objective 
%that depends on some \emph{inner maximization} performed with a different objective function.
%Bilevel programming is a natural tool in our setting, but is intractable in general; our 
%approximation \eqref{eq:loss-full} can be thought of as a tractable relaxation of the 
%bilevel programming objective.

A striking recent security vulnerability discovered in machine learning systems 
is \emph{adversarial test images} that can fool image classifiers 
despite being imperceptible from normal images 
\citep{szegedy2014intriguing,goodfellow2015explaining,carlini2016hidden,
kurakin2016adversarial,papernot2016transferability}. 
These images exhibit vulnerabilities at test time, whereas data poisoning is 
a vulnerability at training time.
However, recent adversarial attacks on reinforcement learners 
\citep{huang2017adversarial,behzadan2017vulnerability,lin2017tactics} 
do blend train and test vulnerabilities. 
A common defense against adversarial test examples is \emph{adversarial training} 
\citep{goodfellow2015explaining}, 
which alters the training objective to encourage robustness.

We note that generative adversarial networks \citep{goodfellow2014gan}, despite their name, 
are not focused on security but rather provide a game-theoretic objective for training generative models.  
%though that work considers a different attack model than the standard data poisoning model.
%\todo{check this}
%\todo{should we move this last part in to data poisoning work? this sounds a bit lame 
%as is}
%\todo{try to cut another line}

Finally, a number of authors have studied the theoretical question of learning in the presence 
of adversarial errors, under a priori distributional assumptions on the data. 
Robust algorithms have been exhibited for mean and covariance estimation and 
clustering \citep{diakonikolas2016robust,lai2016agnostic,charikar2017learning}, 
classification \citep{klivans2009learning,awasthi2014power}, regression 
\citep{nasrabadi2011robust,nguyen2013exact,chen2013robust,bhatia2015robust}
and crowdsourced data aggregation \citep{steinhardt2016avoiding}. 
However, these bounds only hold for specific (sometimes quite sophisticated)
algorithms and are focused on good asymptotic performance, rather than on giving 
good numerical error guarantees for concrete datasets/defenses.
%\todo{okay to not elaborate further?}
%While this work also has the potential to certify defense strategies, it relies on a priori 
%distributional assumptions which may not hold, and requires using specific 
%(sometimes computationally expensive) learning algorithms. In contrast, our certification 
%tool can be applied on any dataset and certifies natural defense algorithms based on 
%data sanitization, and comes coupled with a candidate attack.

\section{Discussion}
\label{sec:discussion}

% What points to make?
% -re-iterate main contribution: framework for studying defense (not just empirically)
In this paper we have presented a tool for studying data poisoning 
defenses that goes beyond empirical validation by providing certificates 
against a large family of attacks modulo the approximations from Section~\ref{sec:attack-defense}.
We stress that our bounds are meant to be used as a way to assess defense strategies in the 
design stage, rather than guaranteeing performance of a deployed learning algorithm (since our 
method needs to be run on the clean data, which we presumably would not have access to at 
deployment time). For instance, if we want to build robust defenses for image 
classifiers, we can assess the performance against attacks on a number of known image 
datasets, in order to gain more confidence in the robustness of the system that we actually 
deploy.
%We saw that while data-dependent defenses 
%can potentially be powerful (by adapting to the structure of the distribution in question) 
%this power adds additional vulnerabilities.
% -further experiments / extensions
%   -other types of classifiers
%   -multi-class
%   -more general defenses; [SKIP?] (cite outlier literature, PCA subspace)
%   [SKIP?] -structured outputs
%   -try to give general flavor that this work readily extends out in new directions

Having applied our framework to binary SVMs, there are a number of 
extensions we can consider: e.g., to other loss functions or 
to multiclass classification. We can also consider defenses beyond the sphere 
and slab constraints considered here---for instance, 
sanitizing text data using a language model, or using
the covariance structure of the clean data \citep{lakhina2004diagnosing}.
The main requirement of our framework is the ability to efficiently maximize
$\ell(\theta; x,y)$ over all feasible $x$ and $y$. For margin-based classifiers
such as SVMs and logistic regression, this only requires maximizing a 
linear function over the feasible set, which is often possible (e.g., via dynamic programming)
even for discrete sets.

Our framework currently does not handle non-convex losses: 
while our method might still be meaningful as a way of generating attacks, our upper bounds would no longer be valid. 
The issue is that an attacker could try to thwart the optimization process and cause the defender to end up in a bad local minimum. 
Finding ways to rule this out without relying on convexity would be quite interesting.

Separately, the bound $\bL(\thetaNom) \lessapprox \bM$ 
was useful because $\bM$ admits the natural minimax formulation 
(\ref{eq:M-to-U}), 
but the worst-case $\bL(\thetaNom)$ can be expressed directly as a bilevel optimization problem 
\citep{mei2015teaching}, 
which is intractable in general but admits a number of heuristics \citep{bard1999}.
Bilevel optimization has been considered in the related setting of 
Stackelberg games \citep{bruckner2011stackelberg,bruckner2012static,zhou2016modeling}, 
and is natural to apply here as well.
%\todo{do we need to differentiate more?} 
%\pw{maybe should cite Jerry's work since
%it's the most related to ours in some sense and directly talks about the bilevel problem}

% SKIP -other direction: *fully certified* defense
% SKIP  -what would we need? handle points of slippage more formally
% SKIP   -true vs. poisoned
% SKIP   -train vs. test
% SKIP   -all clean vs. filtered clean
% SKIP [maybe] call for building comprehensive evaluation suite?

To conclude, we quote \citeauthor{biggio2014security}, who call for the following methodology for 
evaluating defenses:
\newenvironment{myquote}%
  {\list{}{\leftmargin=16pt\rightmargin=16pt\topsep=0pt\parsep=0pt\partopsep=0pt}\item[]}%
  {\endlist}
\begin{myquote}
To pursue security in the context of an arms race it is not sufficient to
\emph{react} to observed attacks, but it is also necessary to
\emph{proactively anticipate} the adversary by \emph{predicting}
the most relevant, potential attacks through a what-if analysis;
this allows one to develop suitable countermeasures \emph{before}
the attack actually occurs, according to the principle of
\emph{security~by~design}.
\end{myquote}
The existing paradigm for such proactive anticipation is to
design various hypothetical attacks against which to test the
defenses. However, such an evaluation is fundamentally limited
because it leaves open the possibility that there is a more
clever attack that we failed to think of. Our approach provides 
a first step towards surpassing this limitation, by not just 
anticipating but certifying the reliability of a defender, thus 
implicitly considering an infinite number of attacks before they occur.

\paragraph{Reproducibility.}
The code and data for replicating our experiments is available
on GitHub (\url{http://bit.ly/gt-datapois})
and Codalab Worksheets (\url{http://bit.ly/cl-datapois}).

\paragraph{Acknowledgments.}
JS was supported by a Fannie \& John Hertz Foundation Fellowship and an NSF Graduate
Research Fellowship.  This work was also partially supported by a Future of Life Institute grant and a grant from the Open Philanthropy Project.
We are grateful to Daniel Selsam, Zhenghao Chen, and Nike Sun, as well as to the 
anonymous reviewers, for a great deal of helpful feedback.

{
%\small
%\setlength{\baselineskip}{2pt}
%\setlength{\bibsep}{2pt plus 2ex}
\bibliographystyle{plainnat}
\bibliography{refdb/all}

\begin{thebibliography}{65}
\providecommand{\natexlab}[1]{#1}
\providecommand{\url}[1]{\texttt{#1}}
\expandafter\ifx\csname urlstyle\endcsname\relax
  \providecommand{\doi}[1]{doi: #1}\else
  \providecommand{\doi}{doi: \begingroup \urlstyle{rm}\Url}\fi

\bibitem[Awasthi et~al.(2014)Awasthi, Balcan, and Long]{awasthi2014power}
P.~Awasthi, M.~F. Balcan, and P.~M. Long.
\newblock The power of localization for efficiently learning linear separators
  with noise.
\newblock In \emph{Symposium on Theory of Computing (STOC)}, pages 449--458,
  2014.

\bibitem[Bard(1999)]{bard1999}
J.~F. Bard.
\newblock \emph{Practical Bilevel Optimization: Algorithms and Applications}.
\newblock Springer, 1999.

\bibitem[Barreno et~al.(2010)Barreno, Nelson, Joseph, and
  Tygar]{barreno2010security}
M.~Barreno, B.~Nelson, A.~D. Joseph, and J.~D. Tygar.
\newblock The security of machine learning.
\newblock \emph{Machine Learning}, 81\penalty0 (2):\penalty0 121--148, 2010.

\bibitem[Behzadan and Munir(2017)]{behzadan2017vulnerability}
V.~Behzadan and A.~Munir.
\newblock Vulnerability of deep reinforcement learning to policy induction
  attacks.
\newblock \emph{arXiv}, 2017.

\bibitem[Bhatia et~al.(2015)Bhatia, Jain, and Kar]{bhatia2015robust}
K.~Bhatia, P.~Jain, and P.~Kar.
\newblock Robust regression via hard thresholding.
\newblock In \emph{Advances in Neural Information Processing Systems (NIPS)},
  pages 721--729, 2015.

\bibitem[Biggio et~al.(2012)Biggio, Nelson, and Laskov]{biggio2012poisoning}
B.~Biggio, B.~Nelson, and P.~Laskov.
\newblock Poisoning attacks against support vector machines.
\newblock In \emph{International Conference on Machine Learning (ICML)}, pages
  1467--1474, 2012.

\bibitem[Biggio et~al.(2013)Biggio, Pillai, Bul{\`o}, Ariu, Pelillo, and
  Roli]{biggio2013clustering}
B.~Biggio, I.~Pillai, S.~R. Bul{\`o}, D.~Ariu, M.~Pelillo, and F.~Roli.
\newblock Is data clustering in adversarial settings secure?
\newblock In \emph{Workshop on Artificial Intelligence and Security (AISec)},
  2013.

\bibitem[Biggio et~al.(2014{\natexlab{a}})Biggio, Fumera, and
  Roli]{biggio2014security}
B.~Biggio, G.~Fumera, and F.~Roli.
\newblock Security evaluation of pattern classifiers under attack.
\newblock \emph{IEEE Transactions on Knowledge and Data Engineering},
  26\penalty0 (4):\penalty0 984--996, 2014{\natexlab{a}}.

\bibitem[Biggio et~al.(2014{\natexlab{b}})Biggio, Rieck, Ariu, Wressnegger,
  Corona, Giacinto, and Roli]{biggio2014malware}
B.~Biggio, K.~Rieck, D.~Ariu, C.~Wressnegger, I.~Corona, G.~Giacinto, and
  F.~Roli.
\newblock Poisoning behavioral malware clustering.
\newblock In \emph{Workshop on Artificial Intelligence and Security (AISec)},
  2014{\natexlab{b}}.

\bibitem[Biggio et~al.(2014{\natexlab{c}})Biggio, Rota, Ignazio, Michele,
  Zemene, Marcello, and Fabio]{biggio2014linkage}
B.~Biggio, B.~S. Rota, P.~Ignazio, M.~Michele, M.~E. Zemene, P.~Marcello, and
  R.~Fabio.
\newblock Poisoning complete-linkage hierarchical clustering.
\newblock In \emph{Workshop on Structural, Syntactic, and Statistical Pattern
  Recognition}, 2014{\natexlab{c}}.

\bibitem[Bishop(2002)]{bishop2002art}
M.~A. Bishop.
\newblock \emph{The art and science of computer security}.
\newblock Addison-Wesley Longman Publishing Co., Inc., 2002.

\bibitem[Br{\"u}ckner and Scheffer(2011)]{bruckner2011stackelberg}
M.~Br{\"u}ckner and T.~Scheffer.
\newblock {S}tackelberg games for adversarial prediction problems.
\newblock In \emph{SIGKDD}, pages 547--555, 2011.

\bibitem[Br{\"u}ckner et~al.(2012)Br{\"u}ckner, Kanzow, and
  Scheffer]{bruckner2012static}
M.~Br{\"u}ckner, C.~Kanzow, and T.~Scheffer.
\newblock Static prediction games for adversarial learning problems.
\newblock \emph{Journal of Machine Learning Research (JMLR)}, 13:\penalty0
  2617--2654, 2012.

\bibitem[Burkard and Lagesse(2017)]{burkard2017analysis}
C.~Burkard and B.~Lagesse.
\newblock Analysis of causative attacks against {SVM}s learning from data
  streams.
\newblock In \emph{International Workshop on Security And Privacy Analytics},
  2017.

\bibitem[Carlini et~al.(2016)Carlini, Mishra, Vaidya, Zhang, Sherr, Shields,
  Wagner, and Zhou]{carlini2016hidden}
N.~Carlini, P.~Mishra, T.~Vaidya, Y.~Zhang, M.~Sherr, C.~Shields, D.~Wagner,
  and W.~Zhou.
\newblock Hidden voice commands.
\newblock In \emph{USENIX Security}, 2016.

\bibitem[Charikar et~al.(2017)Charikar, Steinhardt, and
  Valiant]{charikar2017learning}
M.~Charikar, J.~Steinhardt, and G.~Valiant.
\newblock Learning from untrusted data.
\newblock In \emph{Symposium on Theory of Computing (STOC)}, 2017.

\bibitem[Chen et~al.(2013)Chen, Caramanis, and Mannor]{chen2013robust}
Y.~Chen, C.~Caramanis, and S.~Mannor.
\newblock Robust high dimensional sparse regression and matching pursuit.
\newblock \emph{arXiv}, 2013.

\bibitem[Cretu et~al.(2008)Cretu, Stavrou, Locasto, Stolfo, and
  Keromytis]{cretu2008casting}
G.~F. Cretu, A.~Stavrou, M.~E. Locasto, S.~J. Stolfo, and A.~D. Keromytis.
\newblock Casting out demons: Sanitizing training data for anomaly sensors.
\newblock In \emph{IEEE Symposium on Security and Privacy}, pages 81--95, 2008.

\bibitem[Diakonikolas et~al.(2016)Diakonikolas, Kamath, Kane, Li, Moitra, and
  Stewart]{diakonikolas2016robust}
I.~Diakonikolas, G.~Kamath, D.~Kane, J.~Li, A.~Moitra, and A.~Stewart.
\newblock Robust estimators in high dimensions without the computational
  intractability.
\newblock In \emph{Foundations of Computer Science (FOCS)}, 2016.

\bibitem[Diamond and Boyd(2016)]{diamond2016cvxpy}
S.~Diamond and S.~Boyd.
\newblock {CVXPY}: A {P}ython-embedded modeling language for convex
  optimization.
\newblock \emph{Journal of Machine Learning Research (JMLR)}, 17\penalty0
  (83):\penalty0 1--5, 2016.

\bibitem[Gardiner and Nagaraja(2016)]{gardiner2016security}
J.~Gardiner and S.~Nagaraja.
\newblock On the security of machine learning in malware c\&c detection: A
  survey.
\newblock \emph{ACM Computing Surveys (CSUR)}, 49\penalty0 (3), 2016.

\bibitem[Goodfellow et~al.(2014)Goodfellow, Pouget-Abadie, Mirza, Xu,
  Warde-Farley, Ozair, Courville, and Bengio]{goodfellow2014gan}
I.~J. Goodfellow, J.~Pouget-Abadie, M.~Mirza, B.~Xu, D.~Warde-Farley, S.~Ozair,
  A.~Courville, and Y.~Bengio.
\newblock Generative adversarial nets.
\newblock In \emph{Advances in Neural Information Processing Systems (NIPS)},
  2014.

\bibitem[Goodfellow et~al.(2015)Goodfellow, Shlens, and
  Szegedy]{goodfellow2015explaining}
I.~J. Goodfellow, J.~Shlens, and C.~Szegedy.
\newblock Explaining and harnessing adversarial examples.
\newblock In \emph{International Conference on Learning Representations
  (ICLR)}, 2015.

\bibitem[{{Gurobi {Optimization}, Inc.}}(2016)]{gurobi2016}
{{Gurobi {Optimization}, Inc.}}
\newblock Gurobi optimizer reference manual, 2016.

\bibitem[Huang et~al.(2017)Huang, Papernot, Goodfellow, Duan, and
  Abbeel]{huang2017adversarial}
S.~Huang, N.~Papernot, I.~Goodfellow, Y.~Duan, and P.~Abbeel.
\newblock Adversarial attacks on neural network policies.
\newblock \emph{arXiv}, 2017.

\bibitem[Kakade et~al.(2009)Kakade, Sridharan, and
  Tewari]{kakade2009complexity}
S.~M. Kakade, K.~Sridharan, and A.~Tewari.
\newblock On the complexity of linear prediction: Risk bounds, margin bounds,
  and regularization.
\newblock In \emph{Advances in Neural Information Processing Systems (NIPS)},
  2009.

\bibitem[Kerckhoffs(1883)]{kerckhoffs1883security}
A.~Kerckhoffs.
\newblock La cryptographie militaire.
\newblock \emph{Journal des sciences militaires}, 9, 1883.

\bibitem[Klivans et~al.(2009)Klivans, Long, and Servedio]{klivans2009learning}
A.~R. Klivans, P.~M. Long, and R.~A. Servedio.
\newblock Learning halfspaces with malicious noise.
\newblock \emph{Journal of Machine Learning Research (JMLR)}, 10:\penalty0
  2715--2740, 2009.

\bibitem[Koh and Liang(2017)]{koh2017understanding}
P.~W. Koh and P.~Liang.
\newblock Understanding black-box predictions via influence functions.
\newblock In \emph{International Conference on Machine Learning (ICML)}, 2017.

\bibitem[Kurakin et~al.(2016)Kurakin, Goodfellow, and
  Bengio]{kurakin2016adversarial}
A.~Kurakin, I.~Goodfellow, and S.~Bengio.
\newblock Adversarial examples in the physical world.
\newblock \emph{arXiv}, 2016.

\bibitem[Lai et~al.(2016)Lai, Rao, and Vempala]{lai2016agnostic}
K.~A. Lai, A.~B. Rao, and S.~Vempala.
\newblock Agnostic estimation of mean and covariance.
\newblock In \emph{Foundations of Computer Science (FOCS)}, 2016.

\bibitem[Laishram and Phoha(2016)]{laishram2016curie}
R.~Laishram and V.~V. Phoha.
\newblock Curie: A method for protecting {SVM} classifier from poisoning
  attack.
\newblock \emph{arXiv}, 2016.

\bibitem[Lakhina et~al.(2004)Lakhina, Crovella, and
  Diot]{lakhina2004diagnosing}
A.~Lakhina, M.~Crovella, and C.~Diot.
\newblock Diagnosing network-wide traffic anomalies.
\newblock In \emph{ACM SIGCOMM Computer Communication Review}, volume~34, pages
  219--230, 2004.

\bibitem[Laskov and \v{S}rndi{\`c}(2014)]{laskov2014practical}
P.~Laskov and N.~\v{S}rndi{\`c}.
\newblock Practical evasion of a learning-based classifier: A case study.
\newblock In \emph{Symposium on Security and Privacy}, 2014.

\bibitem[Li et~al.(2016)Li, Wang, Singh, and Vorobeychik]{li2016data}
B.~Li, Y.~Wang, A.~Singh, and Y.~Vorobeychik.
\newblock Data poisoning attacks on factorization-based collaborative
  filtering.
\newblock In \emph{Advances in Neural Information Processing Systems (NIPS)},
  2016.

\bibitem[Lin et~al.(2017)Lin, Hong, Liao, Shih, Liu, and Sun]{lin2017tactics}
Y.~Lin, Z.~Hong, Y.~Liao, M.~Shih, M.~Liu, and M.~Sun.
\newblock Tactics of adversarial attack on deep reinforcement learning agents.
\newblock \emph{arXiv}, 2017.

\bibitem[Liu and Zhu(2016)]{liu2016teaching}
J.~Liu and X.~Zhu.
\newblock The teaching dimension of linear learners.
\newblock \emph{Journal of Machine Learning Research (JMLR)}, 17\penalty0
  (162), 2016.

\bibitem[L{\"{o}}fberg(2004)]{lofberg2004}
J.~L{\"{o}}fberg.
\newblock {YALMIP}: A toolbox for modeling and optimization in {MATLAB}.
\newblock In \emph{CACSD}, 2004.

\bibitem[Maas et~al.(2011)Maas, Daly, Pham, Huang, Ng, and Potts]{maas2011imdb}
A.~L. Maas, R.~E. Daly, P.~T. Pham, D.~Huang, A.~Y. Ng, and C.~Potts.
\newblock Learning word vectors for sentiment analysis.
\newblock In \emph{Association for Computational Linguistics (ACL)}, 2011.

\bibitem[Mei and Zhu(2015{\natexlab{a}})]{mei2015security}
S.~Mei and X.~Zhu.
\newblock The security of latent {D}irichlet allocation.
\newblock In \emph{Artificial Intelligence and Statistics (AISTATS)},
  2015{\natexlab{a}}.

\bibitem[Mei and Zhu(2015{\natexlab{b}})]{mei2015teaching}
S.~Mei and X.~Zhu.
\newblock Using machine teaching to identify optimal training-set attacks on
  machine learners.
\newblock In \emph{Association for the Advancement of Artificial Intelligence
  (AAAI)}, 2015{\natexlab{b}}.

\bibitem[Metsis et~al.(2006)Metsis, Androutsopoulos, and
  Paliouras]{metsis2006spam}
V.~Metsis, I.~Androutsopoulos, and G.~Paliouras.
\newblock Spam filtering with naive {B}ayes -- which naive {B}ayes?
\newblock In \emph{CEAS}, volume~17, pages 28--69, 2006.

\bibitem[Mozaffari-Kermani et~al.(2015)Mozaffari-Kermani, Sur-Kolay,
  Raghunathan, and Jha]{mozaffari2015systematic}
M.~Mozaffari-Kermani, S.~Sur-Kolay, A.~Raghunathan, and N.~K. Jha.
\newblock Systematic poisoning attacks on and defenses for machine learning in
  healthcare.
\newblock \emph{IEEE Journal of Biomedical and Health Informatics}, 19\penalty0
  (6):\penalty0 1893--1905, 2015.

\bibitem[Nasrabadi et~al.(2011)Nasrabadi, Tran, and
  Nguyen]{nasrabadi2011robust}
N.~M. Nasrabadi, T.~D. Tran, and N.~Nguyen.
\newblock Robust lasso with missing and grossly corrupted observations.
\newblock In \emph{Advances in Neural Information Processing Systems (NIPS)},
  2011.

\bibitem[Newell et~al.(2014)Newell, Potharaju, Xiang, and
  Nita-Rotaru]{newell2014practicality}
A.~Newell, R.~Potharaju, L.~Xiang, and C.~Nita-Rotaru.
\newblock On the practicality of integrity attacks on document-level sentiment
  analysis.
\newblock In \emph{Workshop on Artificial Intelligence and Security (AISec)},
  pages 83--93, 2014.

\bibitem[Newsome et~al.(2006)Newsome, Karp, and Song]{newsome2006paragraph}
J.~Newsome, B.~Karp, and D.~Song.
\newblock Paragraph: Thwarting signature learning by training maliciously.
\newblock In \emph{International Workshop on Recent Advances in Intrusion
  Detection}, 2006.

\bibitem[Nguyen and Tran(2013)]{nguyen2013exact}
N.~H. Nguyen and T.~D. Tran.
\newblock Exact recoverability from dense corrupted observations via
  $\ell_1$-minimization.
\newblock \emph{IEEE Transactions on Information Theory}, 59\penalty0
  (4):\penalty0 2017--2035, 2013.

\bibitem[Papernot et~al.(2016{\natexlab{a}})Papernot, McDaniel, and
  Goodfellow]{papernot2016transferability}
N.~Papernot, P.~McDaniel, and I.~Goodfellow.
\newblock Transferability in machine learning: from phenomena to black-box
  attacks using adversarial samples.
\newblock \emph{arXiv}, 2016{\natexlab{a}}.

\bibitem[Papernot et~al.(2016{\natexlab{b}})Papernot, McDaniel, Sinha, and
  Wellman]{papernot2016towards}
N.~Papernot, P.~McDaniel, A.~Sinha, and M.~Wellman.
\newblock Towards the science of security and privacy in machine learning.
\newblock \emph{arXiv}, 2016{\natexlab{b}}.

\bibitem[Park et~al.(2017)Park, Weimer, and Lee]{park2017resilient}
S.~Park, J.~Weimer, and I.~Lee.
\newblock Resilient linear classification: an approach to deal with attacks on
  training data.
\newblock In \emph{International Conference on Cyber-Physical Systems}, pages
  155--164, 2017.

\bibitem[Rubinstein et~al.(2009)Rubinstein, Nelson, Huang, Joseph, Lau, Rao,
  Taft, and Tygar]{rubinstein2009antidote}
B.~Rubinstein, B.~Nelson, L.~Huang, A.~D. Joseph, S.~Lau, S.~Rao, N.~Taft, and
  J.~Tygar.
\newblock Antidote: Understanding and defending against poisoning of anomaly
  detectors.
\newblock In \emph{ACM SIGCOMM Conference on Internet measurement conference},
  2009.

\bibitem[Shalev-Shwartz(2011)]{shalev2011online}
S.~Shalev-Shwartz.
\newblock Online learning and online convex optimization.
\newblock \emph{Foundations and Trends in Machine Learning}, 4\penalty0
  (2):\penalty0 107--194, 2011.

\bibitem[Steinhardt et~al.(2014)Steinhardt, Wager, and
  Liang]{steinhardt2014sparse}
J.~Steinhardt, S.~Wager, and P.~Liang.
\newblock The statistics of streaming sparse regression.
\newblock \emph{arXiv preprint arXiv:1412.4182}, 2014.

\bibitem[Steinhardt et~al.(2016)Steinhardt, Valiant, and
  Charikar]{steinhardt2016avoiding}
J.~Steinhardt, G.~Valiant, and M.~Charikar.
\newblock Avoiding imposters and delinquents: Adversarial crowdsourcing and
  peer prediction.
\newblock In \emph{Advances in Neural Information Processing Systems (NIPS)},
  2016.

\bibitem[Sturm(1999)]{sturm1999guide}
J.~F. Sturm.
\newblock Using {SeDuMi} 1.02, a {MATLAB} toolbox for optimization over
  symmetric cones.
\newblock \emph{Optimization Methods and Software}, 11:\penalty0 625--653,
  1999.

\bibitem[Szegedy et~al.(2014)Szegedy, Zaremba, Sutskever, Bruna, Erhan,
  Goodfellow, and Fergus]{szegedy2014intriguing}
C.~Szegedy, W.~Zaremba, I.~Sutskever, J.~Bruna, D.~Erhan, I.~Goodfellow, and
  R.~Fergus.
\newblock Intriguing properties of neural networks.
\newblock In \emph{International Conference on Learning Representations
  (ICLR)}, 2014.

\bibitem[Tram{\`e}r et~al.(2016)Tram{\`e}r, Zhang, Juels, Reiter, and
  Ristenpart]{tramer2016stealing}
F.~Tram{\`e}r, F.~Zhang, A.~Juels, M.~K. Reiter, and T.~Ristenpart.
\newblock Stealing machine learning models via prediction {API}s.
\newblock In \emph{USENIX Security}, 2016.

\bibitem[Vuurens et~al.(2011)Vuurens, de~Vries, and Eickhoff]{vuurens2011spam}
J.~Vuurens, A.~P. de~Vries, and C.~Eickhoff.
\newblock How much spam can you take? {A}n analysis of crowdsourcing results to
  increase accuracy.
\newblock \emph{ACM SIGIR Workshop on Crowdsourcing for Information Retrieval},
  2011.

\bibitem[Wang(2016)]{wang2016combating}
G.~Wang.
\newblock \emph{Combating Attacks and Abuse in Large Online Communities}.
\newblock PhD thesis, University of California Santa Barbara, 2016.

\bibitem[Xiao et~al.(2012)Xiao, Xiao, and Eckert]{xiao2012adversarial}
H.~Xiao, H.~Xiao, and C.~Eckert.
\newblock Adversarial label flips attack on support vector machines.
\newblock In \emph{European Conference on Artificial Intelligence}, 2012.

\bibitem[Xiao et~al.(2015{\natexlab{a}})Xiao, Biggio, Brown, Fumera, Eckert,
  and Roli]{xiao2015lasso}
H.~Xiao, B.~Biggio, G.~Brown, G.~Fumera, C.~Eckert, and F.~Roli.
\newblock Is feature selection secure against training data poisoning?
\newblock In \emph{International Conference on Machine Learning (ICML)},
  2015{\natexlab{a}}.

\bibitem[Xiao et~al.(2015{\natexlab{b}})Xiao, Biggio, Nelson, Xiao, Eckert, and
  Roli]{xiao2015contamination}
H.~Xiao, B.~Biggio, B.~Nelson, H.~Xiao, C.~Eckert, and F.~Roli.
\newblock Support vector machines under adversarial label contamination.
\newblock \emph{Neurocomputing}, 160:\penalty0 53--62, 2015{\natexlab{b}}.

\bibitem[Xiao(2010)]{xiao2010rda}
L.~Xiao.
\newblock Dual averaging methods for regularized stochastic learning and online
  optimization.
\newblock \emph{Journal of Machine Learning Research (JMLR)}, 11:\penalty0
  2543--2596, 2010.

\bibitem[Yang et~al.(2017)Yang, Wu, Li, and Chen]{yang2017generative}
C.~Yang, Q.~Wu, H.~Li, and Y.~Chen.
\newblock Generative poisoning attack method against neural networks.
\newblock \emph{arXiv}, 2017.

\bibitem[Zhou and Kantarcioglu(2016)]{zhou2016modeling}
Y.~Zhou and M.~Kantarcioglu.
\newblock Modeling adversarial learning as nested {S}tackelberg games.
\newblock In \emph{Pacific-Asia Conference on Knowledge Discovery and Data
  Mining}, 2016.

\end{thebibliography}
}

%\newpage

\appendix

\section{Proof of Proposition~\ref{prop:regret}}
\label{sec:regret-proof}

Proposition~\ref{prop:regret} follows by standard duality arguments which we reproduce 
here. First recall the definition of $\Regret$: for a sequence of loss functions 
$f_t(\theta)$, $t = 1,\ldots, T$, and an algorithm with iterates 
$\theta^{(1)}, \ldots, \theta^{(T)}$, regret is defined as
\begin{equation}
\label{eq:regret-def}
\Regret(T) \eqdef \sum_{t=1}^T f_t(\theta^{(t)}) - \min_{\theta \in \Theta} \sum_{t=1}^T f_t(\theta).
\end{equation}
In our particular case we take 
$f_t(\theta) = \frac{1}{n} L(\theta; \sDc) + \epsilon \ell(\theta; x^{(t+1)}, y^{(t+1)})$. Hence 
\begin{align}
f_t(\theta^{(t)}) 
\nonumber  &= \frac{1}{n} L(\theta^{(t)}; \sDc) + \epsilon \ell(\theta^{(t)}; x^{(t+1)}, y^{(t+1)}) \\ 
  &= \frac{1}{n} L(\theta^{(t)}; \sDc) + \epsilon \max_{(x,y) \in \sF} \ell(\theta^{(t)}; x,y) = U(\theta^{(t)}).
\end{align}
Substituting into \eqref{eq:regret-def} and averaging over $T$, we have
\begin{equation}
\frac{\Regret(T)}{T} = \frac{1}{T}\sum_{t=1}^T U(\theta^{(t)}) - \min_{\theta \in \Theta} \p{\frac{1}{n} L(\theta; \sDc) + \frac{\epsilon}{T} \sum_{t=1}^T \ell(\theta; x^{(t)}, y^{(t)})}.
\end{equation}
For $t = \epsilon n$ the right-hand term is equal to 
$\frac{1}{n} L(\theta; \sDc \cap \{(x^{(t)}, y^{(t)})\}_{t=1}^{\epsilon n})$. 
Letting $\sDp = \{(x^{(t)}, y^{(t)}\}_{t=1}^{\epsilon n}$ and upper-bounding the min over 
$\theta$ by the value at $\thetaTil$, we obtain
\begin{equation}
\frac{1}{n} L(\thetaTil; \sDcp) \geq \frac{1}{T} \sum_{t=1}^T U(\theta^{(t)}) - \frac{\Regret{T}}{T},
\end{equation}
and in particular $\frac{1}{n} L(\thetaTil; \sDcp) \geq U^* - \frac{\Regret(T)}{T}$, as was to be shown.

\section{Defending Against Overfitting Attacks}
\label{sec:overfitting-defense}

In Section~\ref{sec:attack-defense} we claimed that it is possible to defend against 
overfitting attacks with appropriate regularization. In this section we justify this 
claim. The key is the classical theory of \emph{uniform convergence}, which allows 
us to say that, with probability $1-\delta$, the following uniform bound holds:
\begin{equation}
\label{eq:uniform-convergence}
\Big|\frac{1}{N} \sum_{(x,y) \in \sD_c} \ell(\theta; x,y) - \bE_{x,y \sim p^*}[\ell(\theta; x,y)]\Big| \leq E(N, \rho, \delta),
\end{equation}
where $E$ is an error bound that is roughly $\rho \sqrt{\frac{\log(1/\delta)}{N}}$.
More precisely, \citet{kakade2009complexity} show the following:
\begin{theorem}[Corollary 5 of \citet{kakade2009complexity}]
Let $\ell(\theta; x,y)$ be any margin-based loss: 
$\ell(\theta; x,y) = \phi(y\langle \theta, x \rangle)$, where 
$\phi$ is $1$-Lipschitz. Then the bound 
\eqref{eq:uniform-convergence} holds with probability $1-\delta$, 
for $E(N, \rho, \delta) = \rho R \p{\sqrt{\frac{4}{n}} + \sqrt{\frac{\log(1/\delta)}{2n}}}$, 
where $R$ is such that $\|x\|_2 \leq R$ with probability $1$.
\end{theorem}
By setting $\rho$ appropriately relative to $R$ and $n$ we can therefore guarantee
that the train and test losses in \eqref{eq:uniform-convergence} are close together, 
and therefore rule out any overfitting attack (because any attack that makes the 
test loss high would also have to make the train loss high).

\section{Regret Bound for Adaptive RDA}
\label{sec:regret-rda}

Our optimization algorithm (Algorithm~\ref{alg:main}) is similar in spirit to 
Regularized Dual Averaging \citep{xiao2010rda}, but the known regret bounds for 
RDA do not apply directly because the regularizer is chosen adaptively to ensure 
the norm constraint $\|\theta\|_2 \leq \rho$ holds. In fact, a somewhat different 
analysis is required in this case, closer in spirit to that given by 
\citet{steinhardt2014sparse} for sparse linear regression. While the details 
would take us beyond the scope of this paper, we state the regret bound here:
\begin{theorem}
After $T$ steps of the update in Algorithm~\ref{alg:main}, the regret of 
Algorithm~\ref{alg:main} can be bounded as
\begin{align}
\Regret(T) &\leq \frac{\rho^2}{2\eta} + \sum_{t=1}^T \frac{\|g^{(t)}\|_2^2}{2\lambda_t}.
\end{align}
\end{theorem}
We make two observations: first, since $\lambda_t \geq \frac{1}{\eta}$ necessarily, by 
setting $\eta$ to be on the order of $\frac{1}{\sqrt{T}}$ we can ensure average regret 
$\oo(1/\sqrt{T})$. On the other hand, in many instances $\lambda_t$ will actually 
increase linearly with $t$ 
(in order to enforce the norm constraints $\|\theta\|_2 \leq \rho$) in 
which case the average regret decreases at the faster rate $\oo(\frac{\log(T)}{T})$. 
In either case, the average regret goes to $0$ as $T \to \infty$.

\section{Semidefinite Program for $\Up(\theta)$}
\label{sec:slab-sdp}
\label{sec:sdp}

% TODO: write out slab algorithm pseudocode here

% in previous section: defined feasible set in terms of true means
% however, actually use poisoned means
% can this make a big difference?
% in this section extend constraints to handle poisoned means
Here we elaborate on the semidefinite program for $\Up(\theta)$ that was discussed 
in Section~\ref{sec:data-dependent}. Recall the definition of $\Up(\theta)$:
\begin{equation}
\label{eq:Up-def-app}
\Up(\theta) = \frac{1}{n} L(\sDc) + \epsilon \max_{\support(\pip) \subseteq \sF(\pip)} \bE_{\pip}[\ell(\theta; x,y)].
\end{equation}
Our goal is to solve the maximization over $\pip$ in the special case that 
$\sF$ is defined by the data-dependent sphere and slab defenses (with empirical centroids) and 
$\ell(\theta; x,y) = \max(1 - y\langle \theta, x \rangle, 0)$ is the hinge loss.
First, we argue that the optimal $\pi_p$ without loss of generality is supported on 
at most four points $(x_{a,+},1)$, $(x_{b,+},1)$, $(x_{a,-},-1)$, and $(x_{b,-},-1)$, 
where the $x_a$ points are support vectors and the $x_b$ points are non-support vectors.

Indeed, suppose that there are two distinct support vectors which both lie in the positive class. 
Then replacing them both with their midpoint does not affect either $\sF(\pip)$ or 
$\bE_{\pip}[\ell(\theta; x,y)]$; moreover, since $\sF(\pip)$ is convex for fixed $\pip$ 
both points are still feasible. A similar argument applies to the non-support vectors 
and to the negative class, so that indeed we may assume there are at most the four distinct 
points above in $\support(\pip)$.

Now, let $\pi_{a,+}$, $\pi_{a,-}$, $\pi_{b,+}$, and $\pi_{b,-}$ be the weights of these 
points under $\pip$. Letting $\mu_+$ and $\mu_-$ be the empirical means of 
the positive and negative class over $\sD_c$, and $p_+$ and $p_-$ the empirical probability 
of the two classes, we have the following expression for $\hat{\mu}_y$:
\begin{equation}
\hat{\mu}_y(\pip) = \frac{p_y\mu_y + \pi_{a,y}x_{a,y} + \pi_{b,y}x_{b,y}}{p_y + \pi_{a,y} + \pi_{b,y}}.
\end{equation}
Moreover, the objective $\bE_{\pip}[\ell(\theta; x,y)]$ may be written as
\begin{equation}
\label{eq:objective}
\bE_{\pip}[\ell(\theta; x,y)] = \pi_{a,+}(1 - \langle \theta, x_{a,+} \rangle) + \pi_{a,-}(1 + \langle \theta, x_{a,-} \rangle),
\end{equation}
using the assumption that the $x_a$ are support vectors and the $x_b$ are not.

Now, the sphere and slab constraints may be written as
\begin{align}
|\langle x_{i,y} - \hat{\mu}_y, \hat{\mu}_y - \hat{\mu}_{-y} \rangle| \leq s_y, \\
\langle x_{i,y} - \hat{\mu}_y, x_{i,y} - \hat{\mu}_y \rangle \leq r_y^2
\end{align}
for $i \in \{a,b\}$, $y \in \{+1,-1\}$. We also have the constraints 
\begin{align}
1 - y\langle \theta, x_{a,y} \rangle \geq 0 \\
1 - y\langle \theta, x_{b,y} \rangle \leq 0
\label{eq:constraint-final}
\end{align}
for $y \in \{+1,-1\}$ (encoding the constraints that the $x_a$ are support vectors 
and the $x_b$ are not).

A careful examination reveals that, for fixed $\pi_{\{a,b\},\{+,-\}}$, 
all terms in (\ref{eq:objective}-\ref{eq:constraint-final})
can be written as linear inequality constraints in the inner products between the $7$ vectors 
$x_{a,+}, x_{a,-}, x_{b,+}, x_{b,-}, \mu_+, \mu_-, \theta$.
Therefore, by changing variables to the $7 \times 7$ Gram matrix $G$ among these vectors, 
we can express the maximization over $\pip$ in \eqref{eq:Up-def-app} as a semidefinite program 
over these variables, with equality constraints for the known inner products between 
$\mu_+$, $\mu_-$, and $\theta$.

Moreover, for any matrix $G \succeq 0$ satisfying these 
equality constraints, it is possible to recover vectors 
$x_{a,+}$, $x_{a,-}$, $x_{b,+}$, and $x_{b,-}$ (depending on $\mu_+$, 
$\mu_-$, and $\theta$) whose inner products match the Gram matrix $G$. Precisely, 
if $G = \left[ \begin{array}{cc} G_{11} & G_{12} \\ G_{21} & G_{22} \end{array} \right]$ 
is the Gram matrix (with block $1$ being the $4$ vectors $\{x_{a,+}, x_{a,-}, x_{b,+}, x_{b,-}\}$, and block $2$ being 
the $3$ known vectors $\{\mu_{+}, \mu_{-}, \theta\}$), then for any vectors 
$\{v_{a,+}, v_{a,-}, v_{b,+}, v_{b,-}\}$ orthogonal to the span of $\mu_+$, $\mu_-$, and $\theta$, we can take 
\begin{align}
\left[ \begin{array}{cccc} x_{a,+} & x_{a,-} & x_{b,+} & x_{b,-} \end{array} \right]
 = \left[ \begin{array}{cccc} v_{a,+} & v_{a,-} & v_{b,+} & v_{b,-} \end{array} \right] A + \left[ \begin{array}{ccc} \mu_+ & \mu_- & \theta \end{array} \right] B,
\end{align}
where $A^{\top}A = G_{11} - G_{12}G_{22}^{\dagger}G_{21}$ and $B = G_{22}^{\dagger}G_{21}$, 
and $\dagger$ denotes pseudoinverse. This means that solving the SDP 
allows us to not only compute the optimal objective value, but also to actually recover vectors $x$ realizing it.

To finish, we must handle the fact that the weights $\pi_{\{a,b\},\{+,-\}}$ are not known. 
However, they comprise only a $3$-dimensional parameter space, and hence we can approximate 
the maximum over all $\pi_{\{a,b\},\{+,-\}}$ through Monte Carlo simulation (i.e., 
randomly sample the weights a sufficiently large number of times and take the best).

\end{document}